\documentclass[10pt,twocolumn,letterpaper]{article}

\usepackage{iccv}              

\definecolor{iccvblue}{rgb}{0.21,0.49,0.74}
\usepackage[pagebackref,breaklinks,colorlinks,allcolors=iccvblue]{hyperref}
\usepackage{bm}
\usepackage{multirow}
\usepackage{adjustbox}
\usepackage{makecell}
\renewcommand{\paragraph}[1]{\vspace{1.25mm}\noindent\textbf{#1}}
\usepackage{graphicx}
\usepackage{setspace}
\usepackage{graphicx, amsmath, amssymb, caption, subcaption, multirow, overpic, textpos}
\definecolor{baselinecolor}{gray}{.9}
\usepackage{algorithm}
\usepackage{algpseudocode} 

\def\paperID{13620} 
\def\confName{ICCV}
\def\confYear{2025}

\title{Dataset Ownership Verification for Pre-trained Masked Models}

\author{
 Yuechen Xie\textsuperscript{\rm 1},
 Jie Song\textsuperscript{\rm 1}\thanks{Corresponding author},
 Yicheng Shan\textsuperscript{\rm 2},
 Xiaoyan Zhang\textsuperscript{\rm 1}, \\
 Yuanyu Wan\textsuperscript{\rm 1},
 Shengxuming Zhang\textsuperscript{\rm 1},
 Jiarui Duan\textsuperscript{\rm 1},
 Mingli Song\textsuperscript{\rm 1,3,4} \\[2mm]
 $^1$Zhejiang University, $^2$The University of Sydney\\
 $^3$State Key Laboratory of Blockchain and Security, Zhejiang University \\
 $^4$Hangzhou High-Tech Zone (Binjiang) Institute of Blockchain and Data Security \\[2mm]
 {\tt\small \{xyuechen,sjie,zhang\_xy99,wanyy,zsxm1998,jerryduan,brooksong\}@zju.edu.cn} \\
 \tt\small ysha4092@uni.sydney.edu.au \\
}

\begin{document}
\maketitle
\begin{abstract}
High-quality open-source datasets have emerged as a pivotal catalyst driving the swift advancement of deep learning, while facing the looming threat of potential exploitation.
Protecting these datasets is of paramount importance for the interests of their owners. 
The verification of dataset ownership has evolved into a crucial approach in this domain; however, existing verification techniques are predominantly tailored to supervised models and contrastive pre-trained models, rendering them ill-suited for direct application to the increasingly prevalent masked models.
In this work, we introduce the inaugural methodology addressing this critical, yet unresolved challenge, termed Dataset Ownership Verification for Masked Modeling (DOV4MM). 
The central objective is to ascertain whether a suspicious black-box model has been pre-trained on a particular unlabeled dataset, thereby assisting dataset owners in safeguarding their rights. DOV4MM is grounded in our empirical observation that when a model is pre-trained on the target dataset, the difficulty of reconstructing masked information within the embedding space exhibits a marked contrast to models not pre-trained on that dataset. We validated the efficacy of DOV4MM through ten masked image models on ImageNet-1K and four masked language models on WikiText-103. The results demonstrate that DOV4MM rejects the null hypothesis, with a $p$-value considerably below 0.05, surpassing all prior approaches. Code is available at \url{https://github.com/xieyc99/DOV4MM}.
\end{abstract}    
\section{Introduction}
\label{sec:intro}

High-quality open-source datasets~\cite{deng2009imagenet, krizhevsky2009learning} are the cornerstone of breakthroughs in the field of deep learning, providing valuable resources for researchers and developers worldwide and driving rapid technological progress. Many open-source datasets are explicitly designated for academic use only, with restrictions against unauthorized commercial exploitation. 
Yet, as the value of data continues to surge, these datasets increasingly face the threat of misuse. To preserve the legitimate rights of data proprietors and thwart malicious theft or exploitation, it is imperative to secure the integrity and usage rights of open-source datasets.
\begin{figure}[t]
  \centering
   \includegraphics[width=\linewidth, trim=0.5cm 1.3cm 1cm 0.5cm, clip]{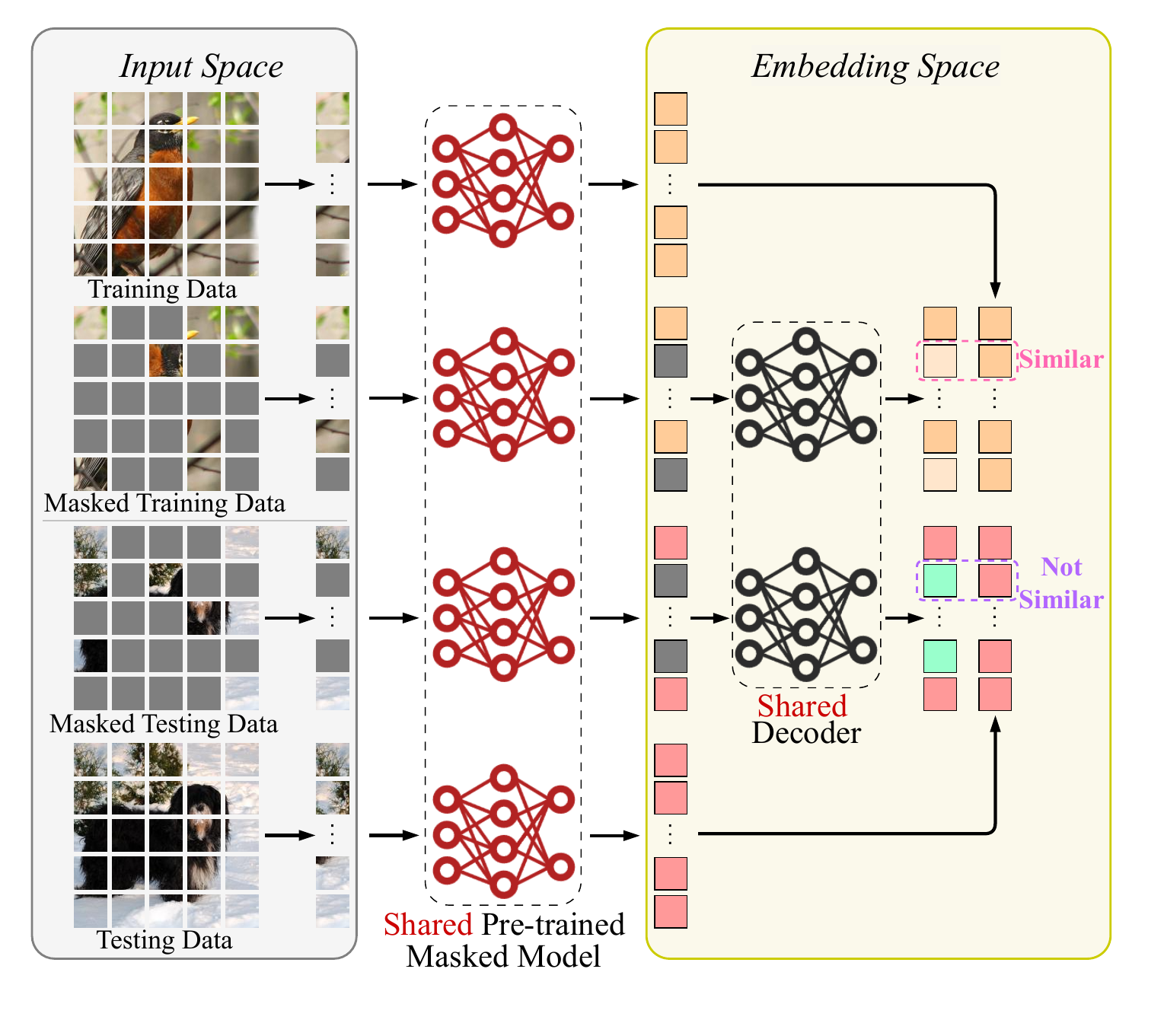}
   \caption{The overview of DOV4MM's motivation. In the embedding space of pre-trained masked models, the reconstruction difficulty of masked information for seen samples in the pre-training phase is lower than that for unseen samples.}
   \vspace{-1.5em}
   \label{fig:motivation}
\end{figure}

Recently, dataset ownership verification (DOV)~\cite{guo2023domain,li2022untargeted,li2023black,guo2024zeromark,xie2025training} has emerged as a defense mechanism to prevent dataset misuse, aiming to protect datasets from being stolen. This technique helps defenders, \textit{i.e.}, dataset owners, detect whether a suspicious black-box model has been trained on their dataset, thus determining if their rights have been infringed. However, most existing DOV methods are designed for supervised learning models~\cite{guo2023domain,li2022untargeted,li2023black,guo2024zeromark}, where verification relies on the relationship between data points and decision boundaries. 
Furthermore, these methods~\cite{guo2023domain,li2022untargeted,li2023black} are heavily reliant on backdoor watermarks. Specifically, models trained on watermarked datasets are embedded with a pre-designed backdoor, allowing defenders to verify dataset ownership by activating the backdoor. However, this strategy not only compromises model performance but also renders the model susceptible to watermark removal techniques~\cite{liu2021wdnet,sun2023defending}.
To overcome these two challenges,~\cite{xie2025dataset} introduced a new DOV technique for contrastive pre-trained models that does not rely on backdoor watermarks, where verification relies on the contrastive relationship gap in the embedding space. Nonetheless, due to the significant differences between the proxy tasks of masked modeling~\cite{he2022masked,zhou2021ibot} and contrastive learning~\cite{chen2020simple,chen2021empirical}, the representations of pre-trained masked models are harder to distinguish~\cite{zhou2022mimco}, leading to less noticeable contrastive relationship gaps in their representations. As a result, the method in~\cite{xie2025dataset} is ineffective for pre-trained masked models, which are widely used in both computer vision~\cite{he2022masked,zhou2021ibot,bao2021beit} and natural language processing~\cite{devlin2018bert,liu2019roberta,lan2019albert}.

In this work, we introduce the pioneering DOV method, termed DOV4MM, designed specifically for pre-trained masked models, to address this important yet unexplored challenge. Notably, DOV4MM operates without relying on backdoor watermarks. It assists defenders in verifying whether a suspicious model has been pre-trained on their public datasets. 
DOV4MM concentrates on the black-box scenario, wherein defenders lack any insight into the model's pre-training configurations (e.g., loss functions and model architecture) and can only interact with the model via Encoder-as-a-Service (EaaS) \cite{sha2023can,liu2022stolenencoder}. It means that defenders can only retrieve feature vectors through the model's API. DOV4MM is grounded in an empirical observation, as depicted in \cref{fig:motivation}. In the embedding space of the pre-trained masked model, the difficulty of reconstructing masked information from seen samples during the pre-training phase is markedly lower than that for unseen samples.

We propose the concept \textit{relative embedding reconstruction difficulty} based on this observation. The defenders can exploit the difference of relative embedding reconstruction difficulty between seen and unseen samples in the suspicious model, thereby determining whether the suspicious model has been pre-trained on their data.  
More specifically, as illustrated in \cref{fig:flowchart}, DOV4MM comprises three key steps:  
(1) Randomly partitioning the public dataset into two disjoint subsets, namely the training dataset and the validation dataset. The decoder is then trained using the training dataset, enabling it to reconstruct masked information in the embedding space;  
(2)  Employing the decoder to compute the embedding reconstruction difficulties of the suspicious model for the training dataset, the validation dataset, and a private dataset (undisclosed to the defender) respectively. These difficulties are then used to calculate the relative embedding reconstruction difficulty; 
(3) Conducting a one-tailed pairwise t-test~\cite{hogg2013introduction} on the relative embedding reconstruction difficulty between the validation dataset and the private dataset, to determine whether the suspicious model has been pre-trained on the defender's public dataset.


In summary, this paper presents three key contributions:
(1) We observe that when a model is pre-trained on a specific target dataset, the difficulty of reconstructing masked information in the embedding space shows significant discrepancies when compared to a model not pre-trained on that dataset;
(2) We introduce the concept of relative embedding reconstruction difficulty and propose a novel DOV technique, DOV4MM. To the best of our knowledge, this is the first DOV technique tailored for pre-trained masked models;
(3) Extensive experiments validate that DOV4MM effectively rejects the null hypothesis, with a $p$-value significantly below 0.05, surpassing all previous methodologies.

\section{Related Work}
\label{sec:related}

\subsection{Data Protection}
\paragraph{Dataset ownership verification.} Dataset ownership verification is an emerging field in data security. Typically, it involves embedding watermarks into the original dataset~\cite{guo2023domain,li2022untargeted,li2023black,tang2023did,guo2024zeromark}. Models trained on the watermarked dataset will be implanted a pre-designed backdoor, allowing defenders to verify dataset ownership simply by triggering it. However, these DOV methods primarily target supervised models, which makes it unable to adapt to the increasingly popular self-supervised models~\cite{chen2020simple,chen2021exploring,chen2021empirical,he2022masked,zhou2021ibot,bao2021beit}.
Recently, backdoor attack methods targeting self-supervised models have been emerging~\cite{li2023embarrassingly,zhang2024data,saha2022backdoor,carlini2021poisoning}, and they are expected to become reliable DOV techniques for self-supervised models. However, like previous backdoor watermarking methods~\cite{guo2023domain,li2022untargeted,li2023black}, they require altering the original dataset's distribution to inject watermarks, which makes it susceptible to various watermark removal mechanisms~\cite{liu2021wdnet,sun2023defending,hayase2021spectre,tejankar2023defending,zheng2024ssl}. A new DOV technique~\cite{xie2025dataset} is proposed for contrastive pre-trained models, which does not rely on backdoor watermarks. However, due to the differing proxy tasks of masked modeling and contrastive learning, the representations learned by pre-trained masked models are more difficult to distinguish~\cite{zhou2022mimco}. Therefore,~\cite{xie2025dataset} cannot be applied to pre-trained masked models directly. To fill this gap, we propose a DOV technique for pre-trained masked models, which also demonstrates that for these models, dataset can be effectively protected without relying on backdoor watermark.

\paragraph{Dataset inference.} Dataset inference~\cite{maini2021dataset} is a state-of-the-art defense approach for preventing model stealing~\cite{sha2023can}. The latest dataset inference method~\cite{dziedzic2022dataset} has expanded its application to self-supervised learning. The intuition behind it is that the log-likelihood of an encoder's output representations is higher on the victim's training data than on test data if it is stolen from the victim. Although it's aimed at encoder theft, it can also be directly used for DOV. However, it requires inferring the entire dataset to model the features of all data. It is prohibitively time-consuming for large datasets, such as ImageNet-1K~\cite{deng2009imagenet}. In contrast, our method achieves accurate verification using only a small fraction of the dataset. For example, we use only 3\% of all training data for accurate verification on ImageNet-1K, which significantly reduces computational cost.

\paragraph{Membership inference.} Membership inference~\cite{shokri2017membership,choquette2021label,carlini2022membership,zhu2024unified} endeavors to determine whether an input was part of the model's training dataset. At present, PartCrop~\cite{zhu2024unified} is a powerful methodology designed for membership inference on visual self-supervised encoder, which takes advantage of the shared part-aware capability among models and stronger part response on the training data. However, it directly relies on the entire high-dimensional representation for membership inference, which contains a large amount of redundant information. In contrast, our method extracts the most critical information for verification from the representations, namely relative embedding reconstruction difficulty, achieving effective verification.


\subsection{Masked Modeling}
Masked modeling~\cite{he2022masked,wei2022masked,devlin2018bert,liu2019roberta} is a self-supervised learning technique which encourages the model to learn useful representations from the unmasked portions of the data. The pre-trained model is applicable to a variety of downstream tasks. Masked modeling techniques originated and developed in natural language processing~\cite{devlin2018bert,liu2019roberta,conneau2019unsupervised}. Due to performance limitations, in the early stages of computer vision, masked modeling had a minimal impact compared to contrastive learning~\cite{chen2020simple,chen2021empirical}. However, once vision Transformers~\cite{dosovitskiy2020image,liu2021swin} surpassed convolutional neural networks~\cite{he2016deep,simonyan2014very} in performance, masked modeling has regained attention as an effective pre-training method for vision Transformers~\cite{li2023masked}, making it one of the mainstream visual self-supervised paradigms alongside contrastive learning. DOV4MM is focused on safeguarding the unlabeled datasets used in pre-trained masked models, ensuring that they are not misused by suspects, thereby securing and fostering the healthy advancement of this field.
\section{Method}
\label{sec:method}

\begin{figure*}[t]
  \centering
   \includegraphics[width=\linewidth, trim=1.1cm 0.7cm 2.2cm 0.6cm, clip]{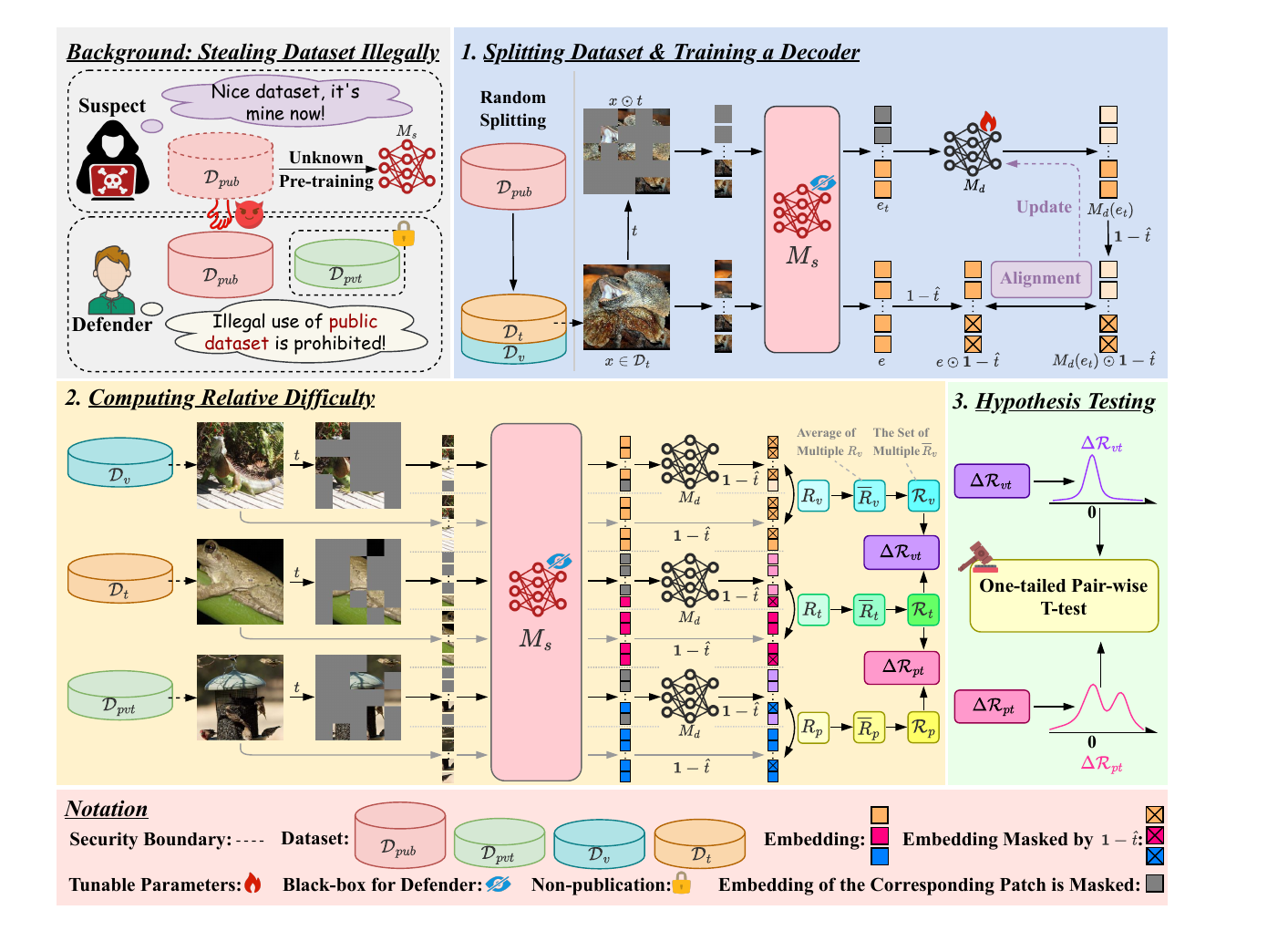}
   \caption{The overview of DOV4MM (best viewed under color conditions).}
   \vspace{-1em}
   \label{fig:flowchart}
\end{figure*}

\subsection{Problem Formulation}
In this study, we focus on the dataset ownership verification problem in black-box scenarios, where two key players are involved: the \textit{defender} and the \textit{suspect}. The defender, assuming the role of the dataset provider, endeavors to ascertain whether the suspicious model, $M_s$, has been unlawfully pre-trained on his public dataset $\mathcal {D}_{pub}$.

$M_s$ can be classified as either illegal or legal based on its training datasets:
(1) \textit{Illegal}: $M_s$ is pre-trained on the data from $\mathcal D_{pub}$, indicating the occurrence of dataset misappropriation;
(2) \textit{Legal}: $M_s$ is pre-trained on an unrelated dataset outside the scope of $\mathcal D_{pub}$, indicating the innocence of the suspect. 
More specifically, for a black-box suspect model, the defender's objective is to determine whether it is illegal or legal based solely on the feature vectors it outputs.

\subsection{Relative Embedding Reconstruction Difficulty}

\subsubsection{Observations and Definitions}

In masked modeling, a key training objective is to hide a portion of the input data (such as patches of an image or words in a text) and train the model to predict the missing parts, thereby learning the underlying structure and useful representations of the data. This training approach leverages the memory capabilities of neural networks, encouraging the model to retain features of the training data. Therefore, we derive the following important insights:

\paragraph{Observation 1.} \textit{In the embedding space of pre-trained masked models, the reconstruction difficulty of masked information for seen samples during pre-training is lower than unseen samples.}
\vspace{0.5em}

We characterize the reconstruction difficulty of masked information in the embedding space as the model's \textit{embedding reconstruction difficulty}. The definition of embedding reconstruction difficulty is as follows:

\paragraph{Definition 1 (Embedding Reconstruction Difficulty).} \textit{Given a pre-trained masked model $M:\mathbb{R}^m\rightarrow\mathbb{R}^n$ and a dataset $\mathcal{D}=\{\bm{x}_i\}_{i=1}^{|\mathcal{D}|}$, where $m$ is the dimension of the input space (\eg, for an image input, $m = C \times W \times H$), $n$ is the dimension of the embedding space, and $|\mathcal{D}|$ is the number of samples in \(\mathcal{D}\). The embedding reconstruction difficulty of $M$ on $\mathcal{D}$ is defined as:}
\begin{equation}
\mathcal{R} = \big\{ \overline{R}_k(\mathcal{D}^k,M) \big| k \in [1,K], \mathcal{D}^k \in \mathcal{D} \big\}
\label{eq_ERD}
\end{equation}
\textit{$K$ is the iterations of sampling, \( \mathcal{D}^k \) is the subset obtained by random sampling from \( \mathcal{D} \) in the \( k \)-th iteration, and \( \overline{R}_k=\frac{1}{|\mathcal{D}^k|}\sum\nolimits_{i = 1}^{|\mathcal{D}^k|}{R_i} \) is the average embedding reconstruction difficulty of \( M \) over all samples in \( \mathcal{D}^k \).}

Given that the difference in embedding reconstruction difficulty between seen and unseen samples during the pre-training phase may not be significant (even though the former is lower than the latter), especially when the reconstruction difficulty for both is relatively high, we introduce the \textit{relative embedding reconstruction difficulty}, a metric that can aid defenders in discerning whether the queried model has been pre-trained on their dataset. The definition of relative embedding reconstruction difficulty is as follows:

\paragraph{Definition 2 (Relative Embedding Reconstruction Difficulty).} \textit{Given the embedding reconstruction difficulties \( \mathcal{R} \) and \( \mathcal{R'} \) of a pre-trained masked model $M$ on datasets \( \mathcal{D} \) and \( \mathcal{D'} \), respectively, the relative embedding reconstruction difficulty of model \( M \) on dataset \( \mathcal{D'} \), with \( \mathcal{D} \) as the standard, is defined as:}
\begin{equation}
\Delta \mathcal{R} = \big\{ \overline{R'}_k - \overline{R}_k \big| k \in [1,K], \overline{R}_k \in \mathcal{R}, \overline{R'}_k \in \mathcal{R'} \big\}
\label{eq_ERRD}
\end{equation}
\textit{$K$ is the iterations of sampling, same as in \cref{eq_ERD}, \( \overline{R}_k \) and \( \overline{R'}_k \) are the \( k \)-th elements of \( \mathcal{R} \) and \( \mathcal{R'} \) respectively.}

\subsubsection{The Calculation of \texorpdfstring{$\Delta \mathcal{R}$}{Lg}}


To compute $\Delta \mathcal{R}$, we first calculate the embedding reconstruction difficulty of the pre-training masked model \( M \) on a single sample \( \bm{x} \). To this end, we introduce two masks \( \bm{t} \) and \( \hat{\bm{t}} \), along with a decoder \( M_d \).

\( \bm{t} \in \{0, 1\}^m \) is a random mask in input space $\mathbb{R}^m$, where the positions corresponding to 0 are masked, and the positions corresponding to 1 are retained. \( \hat{\bm{t}} \in \{0, 1\}^n \) is the mask in embedding space $\mathbb{R}^n$ that corresponds to \( \bm{t} \). Such as, for an image, \( \bm{t} \) masks certain patches in it, while \( \hat{\bm{t}} \) masks the tokens corresponding to the patches masked by \( \bm{t} \).

\( M_d:\mathbb{R}^n\rightarrow\mathbb{R}^n \) is a decoder that can reconstruct the missing information in the embedding space $\mathbb{R}^n$. It uses the output embeddings from the pre-training masked model \( M \) for embedding reconstruction training. Note that \( M_d \) corresponds one-to-one with \( M \). Specifically, a decoder \( M_d \) trained based on \( M \) can effectively reconstruct the masked embeddings of \( M \), but this may not hold for other pre-training masked models.

We use the above \( \bm{t} \), \( \hat{\bm{t}} \), and \( M_d \) to compute the embedding reconstruction difficulty \( R \) of $M$ on a sample $\bm{x}$, as follows:
\begin{equation}
R(\bm{x}, \bm{t}, \hat{\bm{t}}, M, M_d) = \frac{\big\|\big[M_d(\bm{e_t} ) - \bm{e}\big] \odot (\bm{1}-\hat{\bm{t}})\big\|_2^2}{\big\|\bm{1}-\hat{\bm{t}}\big\|_1}
\label{eq_loss}
\end{equation}
where \(\odot\) is the element-wise multiplication, $\bm{e}=M(\bm{x})$, $\bm{e_t}=M(\bm{x} \odot \bm{t})$, $\| \cdot \|_1$ and $\| \cdot \|_2$ are the L1 norm and L2 norm respectively. $\bm{e}$ and $\bm{e_t}$ represent the embeddings obtained from the original sample \( \bm{x} \) and the masked sample \( \bm{x} \odot \bm{t} \) after inputting them into the pre-training masked model \( M \), respectively. In addition, for the reconstruction difficulty of the entire embedding \( M_d(\bm{e_t}) - \bm{e} \), we only consider the reconstruction difficulty at the masked positions \( \bm{1}-\hat{\bm{t}} \) when computing \( R \), as it reflects the reconstruction difficulty of the missing information.
Based on the embedding reconstruction difficulty $R$ on a single sample, we can further calculate \( \mathcal{R} \) and \( \Delta \mathcal{R} \) using \cref{eq_ERD} and \cref{eq_ERRD}.

\subsection{The Proposed DOV4MM}
We propose a method called DOV4MM to verify dataset ownership by relative embedding reconstruction difficulty, as shown in \cref{fig:flowchart}, which comprises three key steps:

(1) \textbf{Splitting Dataset \& Training a Decoder}: Randomly split the public dataset \( \mathcal{D}_{pub} \), the dataset that needs to be protected, into two subsets, referred to as the training dataset \( \mathcal{D}_t \) and the validation dataset \( \mathcal{D}_v \). Then use the suspicious model \( M_s \) and \( \mathcal{D}_t \) to train a decoder \( M_d \), whose training objective is to reconstruct the embedding information lost due to the mask $\bm{t}$;

(2) \textbf{Computing Relative Difficulty}: Perform $K$ samplings on \( \mathcal{D}_t \), \( \mathcal{D}_v \), and \( \mathcal{D}_{pvt} \) (a defender's private dataset which isn't publicly available, and \(M_s\) has not been pre-trained on it), with $N$ samples per sampling. Based on these samples, use \cref{eq_ERD} and \cref{eq_loss} to obtain the embedding reconstruction difficulties \( \mathcal{R}_t \), \( \mathcal{R}_v \), and \( \mathcal{R}_p \) for the suspicious model \(M_s\) on these three datasets. Then, according to \cref{eq_ERRD}, with \( \mathcal{D}_{t} \) as the standard, calculate \( \Delta \mathcal{R}_{vt} \) using \( \mathcal{R}_t \) and \( \mathcal{R}_v \), and calculate \( \Delta \mathcal{R}_{pt} \) using \( \mathcal{R}_t \) and \( \mathcal{R}_p \);

(3) \textbf{Hypothesis Testing}: Conduct a one-sided pair-wise t-test~\cite{hogg2013introduction} on \( \Delta \mathcal{R}_{vt} \) and \( \Delta \mathcal{R}_{pt} \). The null hypothesis, \(H_0\), posits that the mean difference between the paired samples in \( \Delta \mathcal{R}_{pt} \) and \( \Delta \mathcal{R}_{vt} \) is less than or equal to 0, while the alternative hypothesis, denoted as \(H_1\), posits that the mean difference between the paired samples in \( \Delta \mathcal{R}_{pt} \) and \( \Delta \mathcal{R}_{vt} \) is greater than 0. 

If the $p$-value is less than 0.05, we can reject the null hypothesis and conclude that the mean difference between the paired samples in \( \Delta \mathcal{R}_{pt} \) and \( \Delta \mathcal{R}_{vt} \) is greater than 0, \textit{i.e.}, the suspicious model $M_s$ is very likely to have been pre-trained on the public dataset \( \mathcal{D}_{pub} \). Therefore, we can infer that $M_s$ is illegal and \( \mathcal{D}_{pub} \) has been stolen. On the other hand, if the null hypothesis can't be rejected, we think that the mean difference between the paired samples in \( \Delta \mathcal{R}_{pt} \) and \( \Delta \mathcal{R}_{vt} \) is less than or equal to 0, which implies $M_s$ is legal and the suspect is innocent.

\section{Experiments}
\label{sec:exp}

Here, we present the results of DOV4MM on ImageNet-1K~\cite{deng2009imagenet} and WikiText-103~\cite{merity2016pointer}, ablation experiments, and interference resistance analysis. More details and results (hypothesis testing, visualizations, efficiency analysis, etc) can be found in the supplementary materials.

\begin{figure*}[ht]
    \centering
    \includegraphics[width=\textwidth, trim=7.7cm 0.2cm 7.7cm 12.0cm, clip]{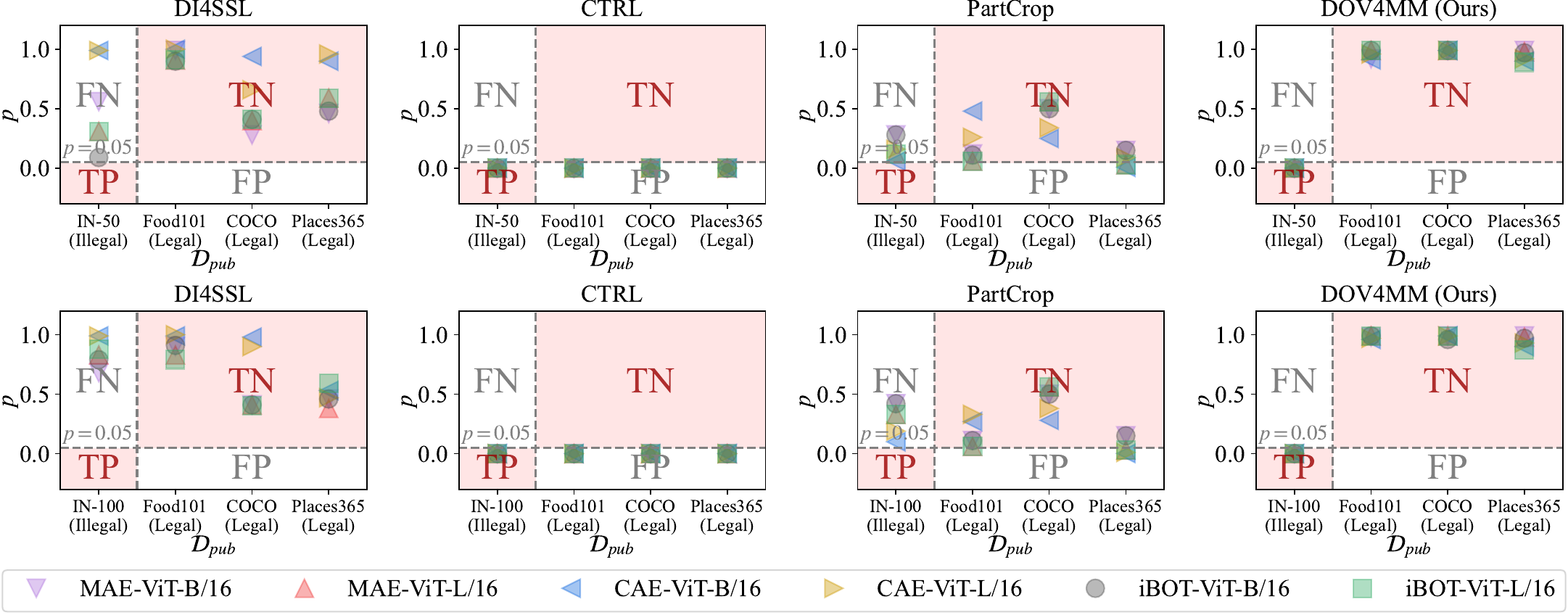}
    \caption{Results of four methods on ImageNet-50 (the first line) and ImageNet-100 (the second line). The pre-training dataset of the suspicious model is ImageNet-50 and ImageNet-100 in two cases respectively. Each pattern represents a suspicious model pre-trained using a specific architecture, masked modeling method, and dataset. ``MAE-ViT-B/16'' represents ViT-B/16 pre-trained using MAE, and the rest follows similarly. On the x-axis, ``IN-50'' and ``IN-100'' represent ImageNet-50 and ImageNet-100, respectively. Additionally, the terms ``Illegal/Legal'' in parentheses on the x-axis indicate the type of the suspicious model in each scenario. We consider illegal/legal models as positive/negative cases and classify each situation based on \(p\)-value.}
    \label{fig:main_exp}
    \vspace{-1em}
\end{figure*}

\subsection{Experimental Setup}

\noindent\textbf{Datasets and models.} We assess the performance of the proposed DOV4MM with ImageNet-1K~\cite{deng2009imagenet} and its subsets, including ImageNet-50 and ImageNet-100. ImageNet-50 is a randomly selected subset of ImageNet-1K, encompassing 50 categories and 63,323 color images. In the same way, ImageNet-100 is also a randomly chosen subset of ImageNet-1K, consisting of 100 categories and 126,532 color images. We conducted the following two main experiments using these datasets:

\noindent\underline{\textbf{Experiments on ImageNet-1K subsets.}} The pre-training dataset of the suspicious model $M_s$ is a ImageNet-1K subset (ImageNet-50 or ImageNet-100). \(\mathcal{D}_{pub}\) is one of the following: the subset of ImageNet-1K ($M_s$'s pre-training dataset), Food101~\cite{bossard2014food}, COCO~\cite{lin2014microsoft}, or Places365~\cite{zhou2017places}. The architecture of $M_s$ includes ViT-B/16~\cite{dosovitskiy2020image} and ViT-L/16~\cite{dosovitskiy2020image}, and the masked modeling methods include MAE~\cite{he2022masked}, CAE~\cite{chen2024context} and iBOT~\cite{zhou2021ibot}.

\noindent\underline{\textbf{Experiments on ImageNet-1K.}} The pre-training dataset of the suspicious model $M_s$ is ImageNet-1K. \(\mathcal{D}_{pub}\) is one of the following: ImageNet-1K ($M_s$'s pre-training dataset), Food101~\cite{bossard2014food}, COCO~\cite{lin2014microsoft}, or Places365~\cite{zhou2017places}. The architecture of the suspicious model $M_s$ includes four architectures (ViT-B/16, ViT-L/16, Swin-B~\cite{liu2021swin}, Swin-L~\cite{liu2021swin}) and ten masked modeling methods (MAE, CAE, iBOT, SimMIM~\cite{xie2022simmim}, MaskFeat~\cite{wei2022masked}, PixMIM~\cite{liu2023pixmim}, BEiT~\cite{bao2021beit}, BEiT v2~\cite{peng2022beit}, MixMAE~\cite{liu2023mixmae} and EVA~\cite{fang2023eva}). Note that the pre-trained models in this experiment are sourced from their official repositories
or MMSelfSup\footnote{\url{https://mmselfsup.readthedocs.io/en/latest/model_zoo.html}}.

When \( \mathcal{D}_{pub} \), the defender's dataset, is the same as the pre-training dataset of the suspicious model \( M_s \), \( M_s \) is deemed illegal; otherwise, \( M_s \) is deemed legal. In all experiments, \( M_s \) is pre-trained using the default settings of each self-supervised method, with 400 epochs and a batch size of 64. The parameter settings for DOV4MM are fixed. Specifically, for convenience, we set \(\mathcal{D}_{pvt}\) as the testing set of \(\mathcal{D}_{pub}\). \( \mathcal{D}_t \) consists of 20,000 randomly selected samples from \( \mathcal{D}_{pub} \), and \( \mathcal{D}_v \) contains the remaining samples from \( \mathcal{D}_{pub} \). The iterations of sampling \( K = 30 \), with \( N = 1,024 \) samples per iteration. The mask strategy for \( \bm{t} \) is random masking, with a masking rate of 75\%. The decoder is a Transformer~\cite{vaswani2017attention} with an embedding dimension of 512, 8 layers, and 16 heads in the multi-head attention mechanism, which is trained for 50 epochs with a batch size of 64, a learning rate of 1e-3, and the Adam optimizer~\cite{kingma2014adam}. More details are provided in the supplementary materials.

\paragraph{Evaluation metrics.}
We classify the suspicious model \( M_s \) as either illegal or legal based on the $p$-value. Specifically, when \( p \) is less than 0.05, we consider that \( M_s \) has been pre-trained on \( \mathcal{D}_{pub} \) and is illegal; when \( p \) is greater than 0.05, \( M_s \) is considered legal. Given that this is a classification task, except for the $p$-value, we also use the sensitivity, specificity and AUROC as the evaluation metrics. Sensitivity is the proportion of correctly predicted positive cases among all actual positive samples, and specificity is the proportion of correctly predicted negative cases among all actual negative samples. They reflect the ability to identify positive and negative samples, respectively.

\subsection{Baselines}

\paragraph{DI4SSL}~\cite{dziedzic2022dataset}: DI4SSL is the most recent method for dataset inference targeting self-supervised models. It can also be applied to dataset ownership verification directly.

\paragraph{CTRL}~\cite{li2023embarrassingly}: CTRL is one of the state-of-the-art backdoor attacks targeting self-supervised models, which can be used for dataset ownership verification through backdoor watermarking. Specifically,we inject the the CTRL trigger as watermark into a small subset of the public dataset. During the verification phase, if the representations of the watermarked images are more similar to each other than those of the non-watermarked images, we can conclude that \( M_s \) was pre-trained using the public dataset.

\paragraph{PartCrop}~\cite{zhu2024unified}: PartCrop is the latest member inference method for self-supervised models.
We make slight modifications to make it applicable for DOV. Specifically, we crop certain parts of objects in both the training and testing images to query their similarity in the embedding space. If higher similarity is observed in the training images, we conclude \( M_s \) has been pre-trained on $\mathcal{D}_{pub}$.


\subsection{Results on ImageNet}

\paragraph{Results on ImageNet-1K subsets.} Our approach is proven effective as illustrated in \cref{fig:main_exp} (refer to supplementary material for specific \(p\)-values), which display the experimental results of baselines and DOV4MM on ImageNet-50 and ImageNet-100. Note that when \(\mathcal{D}_{pub}\) is ImageNet-50 or ImageNet-100, which implies the suspect is illegal (the pre-training dataset of the suspicious model \( M_s \) is ImageNet-50 and ImageNet-100 in two cases respectively), and \(p\)-values should be less than 0.05. However, when \(\mathcal{D}_{pub}\) is Food101, COCO or Places365, the suspect is deemed legal, so \(p\)-values should be greater than 0.05.

\begin{table}
\vspace{0.5em}
\centering
\small
\addtolength{\tabcolsep}{-2.2pt}
\begin{tabular}{ccccc}
\Xhline{1.0pt}
 Dataset&Method&Sensitivity&Specificity&AUROC\\ \hline
 \multirow{4}{*}{ImageNet-50}&DI4SSL& 0.00&\textbf{1.00}&0.50\\
 &CTRL& \textbf{1.00}&0.00& 0.50\\
 &PartCrop& 0.00&0.22& 0.39\\
 &DOV4MM& \textbf{1.00}&\textbf{1.00}& \textbf{1.00}\\ \cline{1-5}
 \multirow{4}{*}{ImageNet-100}&DI4SSL& 0.00&\textbf{1.00}&0.50\\
 &CTRL& \textbf{1.00}&0.00& 0.50\\
 &PartCrop& 0.00&0.28& 0.36\\
 &DOV4MM& \textbf{1.00}&\textbf{1.00}& \textbf{1.00}\\
  \Xhline{1.0pt}
\end{tabular}
\vspace{-1em}
\caption{Sensitivity, specificity, and AUROC of four methods on ImageNet-50 and ImageNet-100. \textbf{Bold} represents the best results.}
\label{tab:main_exp}
\vspace{-1em}
\end{table}

\begin{figure}[ht]
    \centering
    \includegraphics[width=\linewidth, trim=31.1cm 8.8cm 0.2cm 10.6cm, clip]{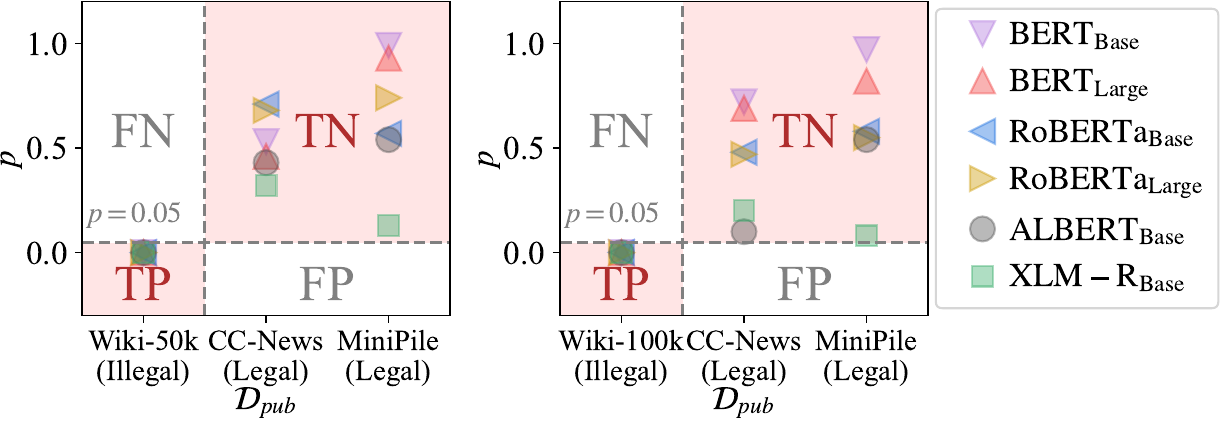}
    \vspace{-2em}
    \caption{The results on WikiText-103 subsets. $M_s$ is pre-trained on WikiText-103-50k (left) and WikiText-103-100k (right) respectively. On the x-axis, ``Wiki-50k'' and ``Wiki-100k'' represent WikiText-103-50k and WikiText-103-100k, respectively. The remaining identifiers are the same as those in \cref{fig:main_exp}.}
    \label{fig:nlp_exp}
    \vspace{-1em}
\end{figure}

As shown in \cref{tab:main_exp}, we calculate sensitivity, specificity and AUROC based on the results in \cref{fig:main_exp}, which demonstrates the superiority of DOV4MM quantitatively. All baselines struggle to accurately distinguish the legality of various scenarios. For DI4SSL and PartCrop, they directly validate high-dimensional representations, while DOV4MM validates the most valuable relative embedding reconstruction difficulty extracted from the representations. CTRL's failure occurs because backdoor watermarks are usually fixed patterns (\eg, fixed-frequency noise for CTRL), designed to ensure the backdoor is successfully embedded into the model. Compared to benign images, the model generates more similar representations for watermarked images, even if the model is pre-trained on a non-watermarked dataset. In contrast, DOV4MM does not rely on backdoor watermarks, thus avoiding this risk.

\paragraph{Results on ImageNet-1K.} We evaluated DOV4MM on ImageNet-1K using ten different masked image modeling (MIM) methods. Note that in this experiment, the pre-training dataset of \( M_s \) is ImageNet-1K, and their pre-trained weight come from the official repository or MMSelfSup. The results of DOV4MM are shown in \cref{tab:in1k_exp}, which demonstrate that even if \( M_s \) is pre-trained on a large-scale dataset like ImageNet-1K, we can still correctly identify malicious behavior, even when using only 3\% of ImageNet-1K data for dataset ownership verification.

\begin{table}
\vspace{0.5em}
\centering
\small
\addtolength{\tabcolsep}{-3.0pt}
\begin{tabular}{cccccc}
\Xhline{1.0pt}
 \multirow{2}{*}{Model}&\multirow{2}{*}{MIM Method}&\multicolumn{4}{c}{\(\mathcal{D}_{pub}\)}\\ \cline{3-6}
 & & IN-1K& Food101& COCO& Places365 \\ \hline
 \multirow{9}{*}{ViT-B/16}& MAE&  $10^{-5}$&  0.99&  0.98&  0.99\\
 & CAE&  $10^{-3}$&  0.99&  0.99&  0.97\\
 & iBOT&  $10^{-3}$&  0.99&  0.99&  0.99\\
 & MaskFeat&  $10^{-4}$&  0.99&  0.99&  0.99\\
 & PixMIM&  $10^{-5}$&  0.99&  0.99&  0.99\\
 & BEiT&  $10^{-3}$&  0.93&  0.99&  0.96\\
 & BEiT v2&  $10^{-5}$&  0.99&  0.99&  0.99\\
 & MixMAE&  $10^{-5}$&  0.95&  0.94&  0.99\\
 & EVA&  $10^{-3}$&  0.92&  0.98&  0.72\\ \hline
 \multirow{3}{*}{ViT-L/16}& MAE&  $10^{-6}$&  0.99&  0.99&  0.99\\
 & CAE&  $10^{-4}$&  0.99&  0.99&  0.99\\
 & iBOT&  $10^{-3}$&  0.99&  0.99&  0.97\\ \hline
 Swin-B& \multirow{2}{*}{SimMIM}&  0.03&  0.99&  0.98&  0.98\\
 Swin-L& &  0.03&  0.99&  0.82&  0.98\\
\Xhline{1.0pt}
\end{tabular}
\vspace{-1em}
\caption{The results (\(p\)-values) of DOV4MM on ImageNet-1K. ``IN-1K'' represents ImageNet-1K. Note that \(p\) should be less than 0.05 in the illegal scenarios and greater than 0.05 in the legal scenarios.}
\label{tab:in1k_exp}
\vspace{-2em}
\end{table}

\subsection{Ablation Studies}

\begin{table*}[t]
\small
\centering
\subfloat[
The width of $M_d$.
\label{tab:width}
]
{
\centering
\begin{minipage}{0.23\linewidth}{\begin{center}
\small
\addtolength{\tabcolsep}{-1.8pt}
\begin{tabular}{cccc}
\Xhline{1.0pt}
Dim & MAE & CAE & iBOT \\
\hline
128 & $10^{-5}$ & $10^{-3}$ & $10^{-3}$ \\
256 & $10^{-6}$ & $10^{-3}$ & $10^{-3}$ \\
512 &  $10^{-5}$ &  $10^{-3}$ &  $10^{-3}$ \\
768 & $10^{-6}$ & 0.01 & $10^{-3}$ \\
1,024 & $10^{-7}$ & 0.01 & $10^{-3}$ \\
\Xhline{1.0pt}
\end{tabular}
\end{center}}\end{minipage}
}
\hspace{0.55em}
\subfloat[
The depth of $M_d$.
\label{tab:depth}
]{
\begin{minipage}{0.23\linewidth}{\begin{center}
\small
\addtolength{\tabcolsep}{-3.0pt}
\begin{tabular}{cccc}
\Xhline{1.0pt}
Blocks & MAE & CAE & iBOT \\
\hline
4 & $10^{-5}$ & $10^{-3}$ & $10^{-3}$ \\
6 & $10^{-6}$ & $10^{-3}$ & $10^{-3}$ \\
8 &  $10^{-5}$ &  $10^{-3}$ &  $10^{-3}$ \\
10 & $10^{-7}$ & $10^{-3}$ & $10^{-3}$ \\
12 & $10^{-6}$ & 0.01 & $10^{-3}$ \\
\Xhline{1.0pt}
\end{tabular}
\end{center}}\end{minipage}
}
\hspace{0.55em}
\subfloat[
The head number of $M_d$.
\label{tab:head_num}
]{
\begin{minipage}{0.23\linewidth}{\begin{center}
\small
\addtolength{\tabcolsep}{-2.2pt}
\begin{tabular}{cccc}
\Xhline{1.0pt}
Heads & MAE & CAE & iBOT \\
\hline
4 & $10^{-6}$ & $10^{-3}$ & $10^{-3}$ \\
8 & $10^{-6}$ & $0.01$ & $10^{-3}$ \\
16 &  $10^{-5}$ &  $10^{-3}$ &  $10^{-3}$ \\
32 & $10^{-6}$ & 0.01 & $10^{-3}$ \\
64 & $10^{-6}$ & $10^{-3}$ & $10^{-3}$ \\
\Xhline{1.0pt}
\end{tabular}
\end{center}}\end{minipage}
}
\hspace{0.55em}
\subfloat[
The size of $\mathcal{D}_{t}$.
\label{tab:n1}
]{
\begin{minipage}{0.23\linewidth}{\begin{center}
\small
\addtolength{\tabcolsep}{-3.2pt}
\begin{tabular}{cccc}
\Xhline{1.0pt}
$|\mathcal{D}_{t}|$ & MAE & CAE & iBOT \\
\hline
10,000 & $10^{-4}$ & 0.02 & 0.01 \\
20,000 &  $10^{-5}$ &  $10^{-3}$ &  $10^{-3}$ \\
30,000 & $10^{-6}$ & $10^{-3}$ & $10^{-3}$ \\
40,000 & $10^{-6}$ & 0.01 & $10^{-3}$ \\
50,000 & $10^{-6}$ & $10^{-3}$ & $10^{-3}$ \\
\Xhline{1.0pt}
\end{tabular}
\end{center}}\end{minipage}
}
\\
\centering
\vspace{.3em}
\subfloat[
Masking ratio.
\label{tab:ratio}
]{
\begin{minipage}{0.23\linewidth}{\begin{center}
\small
\addtolength{\tabcolsep}{-3.7pt}
\begin{tabular}{cccc}
\Xhline{1.0pt}
Ratio (\%) & MAE & CAE & iBOT \\
\hline
30 & $10^{-6}$ & $10^{-3}$ & $10^{-3}$ \\
45 & $10^{-6}$ & $10^{-3}$ & $10^{-3}$ \\
60 & $10^{-6}$ & $10^{-3}$ & $10^{-3}$ \\
75 &  $10^{-5}$ &  $10^{-3}$ &  $10^{-3}$ \\
90 & $10^{-6}$ & 0.04 & $10^{-3}$ \\
\Xhline{1.0pt}
\end{tabular}
\end{center}}\end{minipage}
}
\hspace{0.55em}
\subfloat[
The iterations of sampling.
\label{tab:k}
]{
\begin{minipage}{0.23\linewidth}{\begin{center}
\small
\addtolength{\tabcolsep}{-1.0pt}
\begin{tabular}{cccc}
\Xhline{1.0pt}
$K$ & MAE & CAE & iBOT \\
\hline
10 & $10^{-3}$ & 0.05 & 0.08 \\
20 & $10^{-7}$ & 0.02 & $10^{-3}$ \\
30 &  $10^{-5}$ &  $10^{-3}$ &  $10^{-3}$ \\
40 & $10^{-8}$ & $10^{-3}$ & $10^{-4}$ \\
50 & $10^{-5}$ & $10^{-3}$ & $10^{-4}$ \\
\Xhline{1.0pt}
\end{tabular}
\end{center}}\end{minipage}
}
\hspace{0.55em}
\subfloat[
Sample number per iteration.
\label{tab:n}
]{
\begin{minipage}{0.23\linewidth}{\begin{center}
\small
\addtolength{\tabcolsep}{-2.3pt}
\begin{tabular}{cccc}
\Xhline{1.0pt}
$N$ & MAE & CAE & iBOT \\
\hline
256 & 0.02 & 0.09 & 0.08 \\
512 & $10^{-4}$ & 0.10 & $10^{-3}$ \\
1,024 &  $10^{-5}$ &  $10^{-3}$ &  $10^{-3}$ \\
2,048 & $10^{-10}$ & $10^{-4}$ & $10^{-6}$ \\
4,096 & $10^{-18}$ & $10^{-9}$ & $10^{-10}$ \\
\Xhline{1.0pt}
\end{tabular}
\end{center}}\end{minipage}
}
\hspace{0.6em}
\subfloat[
The object of the t-test.
\label{tab:relative}
]{
\begin{minipage}{0.23\linewidth}{\begin{center}
\small
\addtolength{\tabcolsep}{0.4pt}
\begin{tabular}{cc}
\Xhline{1.0pt}
T-test's Object & SimMIM \\
\hline
$\Delta \mathcal{R}$ &  0.03 \\
$\mathcal{R}$ & 0.10 \\
\Xhline{1.0pt}
\end{tabular}
\vspace{3.7em}
\end{center}}\end{minipage}
}
\vspace{-1em}
\caption{DOV4MM ablation experiments on ImageNet-1K. We report $p$-values. The default settings is: the decoder $M_d$ has depth 8, width 512 and attention heads 16, the size of the training dataset for $M_d$ is 20,000, and the masking ratio is 75\%. Sampling iterations and the number of samples per iteration are 30 and 1,024, respectively. The object of t-test is $\Delta \mathcal{R}$.}
\label{tab:ablations} \vspace{-1.5em}
\end{table*}

The suspicious model $M_s$ used in all ablation experiments is pre-trained on ImageNet-1K, as shown in \cref{tab:ablations}. The model used in MAE, CAE, and iBOT is ViT-B/16, while the model used in SimMIM is Swin-B. Note that in this section, the public dataset \( \mathcal{D}_{pub} \) is ImageNet-1K, so $p$-values need to be below 0.05.

\paragraph{Decoder $M_d$.} We experimented with decoders of varying widths (number of channels), depths (number of Transformer blocks), and the number of attention heads, as shown in \cref{tab:width}, \cref{tab:depth} and \cref{tab:head_num}. The results indicate that the scale of the $M_d$ does not have a significant impact on DOV4MM. Surprisingly, when the $M_d$ is only 3\% of the size of $M_s$ (ViT-B/16), DOV4MM still remains effective.

\paragraph{The size of the training dataset \( \mathcal{D}_t \) for \( M_d \).} We randomly selected different sizes of \( \mathcal{D}_t \) from \( \mathcal{D}_{pub} \) to train \( M_d \), and the results are shown in \cref{tab:n1}. It indicates that the larger \( \mathcal{D}_t \) is, the better the performance of DOV4MM. This is because a very small \( \mathcal{D}_t \) cannot effectively teach \( M_d \) how to reconstruct embeddings. Experiments show that even when we train \( M_d \) with only 10,000 images (0.8\% of the ImageNet-1K), we can achieve correct results.

\paragraph{Masking ratio.} We evaluate DOV4MM at different masking ratios, and the results are shown in \cref{tab:ratio}. DOV4MM is less sensitive to the ratios, and works well across a wide range of masking ratios (30-90\%).

\paragraph{$K$ and $N$.} \cref{tab:k} and \cref{tab:n} show the performance of DOV4MM under different sampling iterations ($K$) and the number of samples per iteration ($N$), which indicates that as $K$ and $N$ increase, the performance of DOV4MM improves. This is because as $K$ and $N$ increase, more data is used in the t-test, making the results more precise. Excitingly, even with just 40,000 images for testing (3\% of the ImageNet-1K), we can achieve satisfactory results.

\paragraph{T-test's object.} We use \( \Delta \mathcal{R} \) or \( \mathcal{R} \) as the object for t-test. As shown in \cref{tab:relative}, using only \( \mathcal{R} \) does not yield correct results ($p$-value exceeds 0.05).

\subsection{Results on Masked Language Modeling}

Masked modeling methods originated in natural language processing (NLP), and we selected four masked language modeling methods, BERT~\cite{devlin2018bert}, RoBERTa~\cite{liu2019roberta}, ALBERT~\cite{lan2019albert} and XLM-R~\cite{conneau2019unsupervised}, to evaluate DOV4MM. The training dataset of the suspicious model $M_s$ is subsets of WikiText-103~\cite{merity2016pointer}, which contains text from English-language wikipedia articles. We denote the subsets as WikiText-103-50k and WikiText-103-100k, which contain 50,000 and 100,000 random samples from WikiText-103, respectively. The training hyperparameters of $M_s$ are provided in the supplementary materials. Note that $M_s$ uses ``[MASK]'' as the mask token for BERT/ALBERT pre-training, and ``\textless mask\textgreater'' for RoBERTa/XLM-R. Since the defender is unaware of $M_s$'s mask token, we default to using ``[MASK]'' as the mask token in DOV4MM. Additionally, apart from a 20\% masking ratio, other parameters are identical to those of visual DOV4MM.

The defender's public dataset \( \mathcal{D}_{pub} \) is one of the following: the subset of WikiText-103 (the pre-training dataset of \( M_s \)), CC-News~\cite{Hamborg2017}, and MiniPile~\cite{kaddour2023minipile}. The results are shown in \cref{fig:nlp_exp}, where DOV4MM can work in various cases, indicating that it can effectively scale to NLP.

\subsection{The Interference Resistance of DOV4MM}

In this section, we study that whether DOV4MM remains effective when facing more covert data theft. More experimental details can be found in the supplementary material.

\begin{table}
\small
\addtolength{\tabcolsep}{1.5pt}
\begin{tabular}{ccccc}

\Xhline{1.0pt}
 MIM Method& Model& w/o es&  w/ es (patience=15) \\ \hline
 \multirow{2}{*}{MAE} &ViT-B/16 &$10^{-5}$ &$10^{-5}$\\
 &ViT-L/16 &$10^{-6}$ &$10^{-4}$ \\
  \Xhline{1.0pt}
\end{tabular}
\vspace{-1em}
\caption{We report the $p$-values of $M_s$, whose pre-training dataset is ImageNet-100, with or without early stopping (es).}
\label{tab:es}
\vspace{-1.5em}
\end{table}

\paragraph{Early stopping.} The suspect may use early stopping to prevent overfitting. Therefore, we tested the performance of DOV4MM under early stopping. $\mathcal{D}_{pub}$ is ImageNet-100, and the $p$-value needs to be less than 0.05. As shown in \cref{tab:es}, DOV4MM remains effective in this case.

\begin{table}
\small
\addtolength{\tabcolsep}{-2.1pt}
\begin{tabular}{cccccc}

\Xhline{1.0pt}
 $\mathcal{D}_{d}$& MIM Method& Model& w/o ft& w/ ft& \textcolor{gray!60}{Acc (\%)}\\ \hline

 \multirow{4}{*}{Food101} &\multirow{2}{*}{MAE} &ViT-B/16 &$10^{-5}$ &$10^{-4}$ &\textcolor{gray!60}{60.65}\\
 & &ViT-L/16 &$10^{-5}$ &$10^{-5}$  &\textcolor{gray!60}{67.79}\\ \cline{2-6}
 & \multirow{2}{*}{iBOT} &ViT-B/16 &$10^{-3}$ &$10^{-3}$ &\textcolor{gray!60}{80.28} \\
 & &ViT-L/16 &$10^{-3}$ &$10^{-3}$ &\textcolor{gray!60}{81.99} \\ \hline
 \multirow{4}{*}{Places365} &\multirow{2}{*}{MAE} &ViT-B/16 &$10^{-5}$ &$10^{-5}$ &\textcolor{gray!60}{42.47} \\
 & &ViT-L/16 &$10^{-5}$ &$10^{-5}$ &\textcolor{gray!60}{45.98} \\ \cline{2-6}
 & \multirow{2}{*}{iBOT} &ViT-B/16 &$10^{-3}$ &$10^{-3}$ &\textcolor{gray!60}{48.08} \\
 & &ViT-L/16 &$10^{-3}$ &0.01 &\textcolor{gray!60}{48.85} \\
  \Xhline{1.0pt}
\end{tabular}
\vspace{-1em}
\caption{We report the $p$-values of \(M_s\), whose pre-training dataset is ImageNet-1K, with or without fine-tuning (ft) on downstream task. \(\mathcal{D}_{d}\) is the downstream dataset, and ``Acc'' is the classification accuracy of the fine-tuned \(M_s\) on the downstream task.}
\label{tab:finetune_down}
\vspace{-1.5em}
\end{table}

\paragraph{Fine-tuning \(M_s\) on downstream task.} Given \(M_s\) pre-trained on ImageNet-1K, we fine-tune it using on Food101 and Places365 for image classification, then DOV4MM performs the test based on the logits output from \(M_s\). Note that in this case, the output of the fine-tuned \(M_s\) is logits, not tokens. Therefore, when calculating the reconstruction difficulty, we no longer use the embedding mask \(\hat{\bm{t}}\), but instead calculate the reconstruction difficulty for the entire logits. Moreover, the decoder is no longer a Transformer but instead a simple three-layer fully connected network. $\mathcal{D}_{pub}$ is ImageNet-1K, and the $p$-value should be less than 0.05. As shown in \cref{tab:finetune_down}, DOV4MM can still work even when defender has no access to the feature vectors.

\vspace{-0.6em}
\section{Conclusion}
\label{sec:conclusion}
\vspace{-0.2em}

In this work, we propose DOV4MM for verifying dataset ownership in masked modeling. Specifically, we propose the concept of relative embedding reconstruction difficulty based on reconstruction characteristics of masked embedding. The experiment demonstrates its effectiveness. Future work includes (1) extending our method to other self-supervised approaches; (2) protecting other types of data; (3) exploring other privacy risks associated with encoders.

\section{Acknowledgment}
This work was partially supported by the Pioneer R\&D Program of Zhejiang (No.2024C01021), and Zhejiang Province High-Level Talents Special Support Program ``Leading Talent of Technological Innovation of Ten-Thousands Talents Program" (No. 2022R52046).
{
    \small
    \bibliographystyle{ieeenat_fullname}
    \bibliography{main}
}

\clearpage
\setcounter{page}{1}
\maketitlesupplementary
\appendix
\setcounter{table}{0}
\renewcommand{\thetable}{S\arabic{table}}

\section{The Details of Experiments}

\subsection{Datasets Used}

\paragraph{ImageNet-1K}~\cite{deng2009imagenet}: A large-scale visual dataset containing over 14 million colored images across 1000 classes. As is commonly done, we resize all images to be of size 224x224. The specific categories we use in ImageNet-50 and ImageNet-100 are listed in ImageNet-50.txt and ImageNet-100.txt in the supplementary material.

\paragraph{Food101}~\cite{bossard2014food}: A dataset contains images of 101 different food categories, with 1,000 images per category, primarily used for food recognition tasks.

\paragraph{COCO}~\cite{lin2014microsoft}: A dataset includes over 328,000 images across 80 object categories, and is widely used for tasks such as object detection, segmentation, and captioning.

\paragraph{Places365}~\cite{zhou2017places}: A dataset contains 365 different scene categories, aimed at improving the accuracy of scene classification and understanding, with over 1.8 million images.

\paragraph{WikiText-103}~\cite{merity2016pointer}: A large-scale dataset derived from Wikipedia articles, containing over 100 million tokens. It is primarily used for language modeling and text generation tasks, focusing on maintaining high-quality, real-world text.

\paragraph{CC-News}~\cite{Hamborg2017}: A dataset consists of news articles collected from the Common Crawl web corpus. It includes over 300 million English news articles, and is commonly used for tasks such as news classification, topic modeling and so on.

\paragraph{MiniPile}~\cite{kaddour2023minipile}: MiniPile is a smaller, curated subset of the Pile dataset~\cite{gao2020pile}, designed for training large-scale language models. It contains diverse text data across multiple domains, such as books, academic papers, and web pages, to improve the generalization of NLP models.

\subsection{Pre-trained MIM Models Used}
The pre-trained models we use on ImageNet-1K all come from the official repository or MMSelfSup, as follows:

\paragraph{MAE}: \url{https://github.com/facebookresearch/mae}.

\paragraph{CAE}: \url{https://github.com/lxtGH/CAE}.

\paragraph{iBOT}: \url{https://github.com/bytedance/ibot}.

\paragraph{SimMIM}: \url{https://github.com/microsoft/SimMIM}.

\paragraph{Other Models}: \url{https://mmselfsup.readthedocs.io/en/latest/model_zoo.html}

\subsection{Parameter Setting Details}
\begin{itemize}[leftmargin=*]
    \item \textbf{Training of masked image modeling (MIM) models}. All MIM models are trained for 400 epochs with a batch size of 64, and the learning rate, masking ratio and optimizer parameters are set to the default settings for each respective method.
    
    \item \textbf{Training of masked language modeling (MLM) models}. All MLM models are trained for 400 epochs with a batch size of 64, a maximum input sequence length of 128, a masking ratio of 20\%, a learning rate of 5e-, an optimizer of Adam, and the mask token is [MASK]. Other parameters are set to the default settings for each respective method.

    \item \textbf{Decoder and its training}. Decoder is a Transformer~\cite{vaswani2017attention} with an embedding dimension of 512, 8 layers, and 16 heads in the multi-head attention mechanism. The training set for the decoder consists of 20,000 randomly sampled examples from the public dataset. The training of the decoder is performed for 50 epochs with a batch size of 64, a learning rate of 1e-3, and the Adam optimizer. The input masking ratio is 75\% (20\% in NLP). The patch size of the mask is $16 \times 16$.

    \item \textbf{Validation phase}. The validation set consists of the remaining portion of the public dataset after excluding the decoder's training set. The sampling frequency is 30, with 1,024 samples per sampling. During inference, the batch size is 256. In the t-test, the significance level \( \alpha \) is 0.05. The ratio and the patch size of the random masking is the same as when training the decoder.

    \item \textbf{Fine-tuning using MIM method}. The hyperparameters used for fine-tuning are the same as those used during pre-training, with the fine-tuning epoch set to 50.

    \item \textbf{Fine-tuning on downstream task}. We add a fully connected layer (classification head) at the output end of the encoder for downstream classification tasks. During fine-tuning, only the classification head is fine-tuned with 50 epochs, a learning rate of 1e-3, the Adam optimizer, and a batch size of 256.
\end{itemize}

\subsection{Hypothesis Testing}
In our experiment, the null hypothesis $H_0$ and the alternative hypothesis $H_1$ for the one-sided pair-wise t-test are as follows:
\begin{itemize}[leftmargin=*]
    \item \textbf{$H_0$}: The mean difference between the paired samples in \( \Delta \mathcal{R}_{pt} \) and \( \Delta \mathcal{R}_{vt} \) is less than or equal to 0.
    
    \item \textbf{$H_1$}: The mean difference between the paired samples in \( \Delta \mathcal{R}_{pt} \) and \( \Delta \mathcal{R}_{vt} \) is greater than 0.
\end{itemize}
Our hypothesis testing can be divided into the following three steps:
\begin{enumerate}[leftmargin=*]
    \item \textbf{Calculate the $t$-statistic.} First, calculate the mean paired differences between the elements in \( \Delta \mathcal{R}_{pt} \) and \( \Delta \mathcal{R}_{vt} \):
    \begin{equation}
    \overline{d} = \frac{1}{K}\sum\nolimits_{k=1}^K (\Delta \mathcal{R}_{pt}^k-\Delta \mathcal{R}_{vt}^k)
    \label{eq_mean}
    \end{equation}
    where $K$ is the iterations of sampling, $\Delta \mathcal{R}_{pt}^k$ and $\Delta \mathcal{R}_{vt}^k$ are the $k$-th elements in \( \Delta \mathcal{R}_{pt} \) and \( \Delta \mathcal{R}_{vt} \), respectively. Then, calculate the standard deviation of the differences:
    \begin{equation}
    s_d = \sqrt{\frac{1}{K-1}\sum\nolimits_{k=1}^K\big(d_k-\overline{d}\big)^2}
    \label{eq_std}
    \end{equation}
    where $d_k=\Delta \mathcal{R}_{pt}^k-\Delta \mathcal{R}_{vt}^k$. Finally, calculate the $t$-statistic:
    \begin{equation}
    t = \frac{\overline{d}}{s_d/\sqrt{K}}
    \label{eq_t}
    \end{equation}
    The $t$-statistic $t$ follows a $t$-distribution with $K-1$ degrees of freedom.
    
    \item \textbf{Calculate the $p$-value.} We first calculate the $p$-value for the two-tailed test:
    \begin{equation}
    p = 2P(T>|t|)
    \end{equation}
    Then, based on the $t$-statistic, the final $p$-value is determined. When the $t$-statistic is greater than 0, $p = p/2$. When the $t$-statistic is less than 0, $p = 1 - p/2$.

    \item \textbf{Determine significance.} Our significance level \(\alpha\) is set to 0.05. If \(p \leq \alpha\), we reject the null hypothesis \(H_0\) and consider the suspicious model is illegal. If \(p > \alpha\), we fail to reject the null hypothesis and consider the suspicious model is legal.
\end{enumerate}

The one-sided pair-wise t-test we use can be easily implemented in Python with just a few lines of code, as shown in \cref{alg}. Here, $t$ is the $t$-statistic and $p$ is the $p$-value. When the output $p$-value is less than 0.05, we conclude that $\Delta \mathcal{R}_{pt}$ is significantly greater than $\Delta \mathcal{R}_{vt}$, meaning $M_s$ illegally used $\mathcal{D}_{pub}$ for training. Conversely, when the $p$-value is greater than 0.05, we conclude that $\Delta \mathcal{R}_{pt}$ is not significantly greater than $\Delta \mathcal{R}_{vt}$, meaning $M_s$ is legal.
\begin{algorithm}[ht]
\begin{algorithmic}[1] 
\caption{One-tailed pair-wise t-test}
\State \textbf{Input}: $\Delta \mathcal{R}_{vt}$, $\Delta \mathcal{R}_{pt}$
\State import scipy
\State $t,p$ = scipy.state.ttest\_ind($\Delta \mathcal{R}_{pt}$, $\Delta \mathcal{R}_{vt}$)
\If{$t > 0$}
    \State $p = p / 2$
\Else
    \State $p = 1 - p / 2$
\EndIf
\State \textbf{Output}: $p$
\label{alg}
\end{algorithmic}
\end{algorithm}

\section{Efficiency Analysis}

We calculated the time required for DOV4MM and three other baselines to perform one validation on ImageNet-1K and Places365. The experiment was conducted using an NVIDIA GeForce RTX 4090. The suspicious model is a ViT-B/16 pre-trained on ImageNet-1K using MAE. Here, the settings of DOV4MM is: the decoder $M_d$ has depth 4, width 128 and attention heads 4, the size of the training dataset for $M_d$ is 10,000. Sampling iterations $K$ and the number of samples per iteration $N$ are 10 and 512, respectively. The training epoch for \(M_d\) is 5. As shown in \cref{tab:time}, only DOV4MM achieved correct results while maintaining a certain level of efficiency. Compared to DI4SSL, we do not need to infer the entire dataset, thus achieving a performance advantage in efficiency. In contrast to CTRL and PartCrop, DOV4MM consumed more time because it requires training a decoder, but it can extract the most valuable relative embedding reconstruction difficulty from redundant representations to obtain correct validation results, which CTRL and PartCrop cannot achieve.

\begin{table}[ht]
\centering
\small
\begin{tabular}{cccc}
\Xhline{1.0pt}
\multirow{2}{*}{Method}& \multirow{2}{*}{$\overline{\tau}$}& \multicolumn{2}{c}{\(\mathcal{D}_{pub}\)} \\  \cline{3-4}
& & ImageNet-1K& Places365 \\  \hline
DI4SSL& 10034s & {1.00} & {0.84} \\
CTRL& 247s & {\(10^{-150}\)} & {\(10^{-140}\)} \\
PartCrop& 128s & {0.71} & {0.77} \\
DOV4MM& 353s & {\(10^{-3}\)} & {0.41} \\
\Xhline{1.0pt}
\end{tabular}
\caption{Efficiency analysis. \( \overline{\tau} \) is the average time taken by each method to perform one validation on ImageNet-1K and Places365. Note that \(p\) should be less than 0.05 in the illegal scenarios and greater than 0.05 in the legal scenarios.}
\label{tab:time}
\vspace{-1.5em}
\end{table}

\section{The Interference Resistance of DOV4MM}

\paragraph{Fine-tuning \(M_s\) using MIM methods.} Given \(M_s\) pre-trained on ImageNet-1K, we fine-tune it using the same MIM method on another dataset, then perform DOV4MM on these fine-tuned models. $\mathcal{D}_{pub}$ is ImageNet-1K, and the $p$-value needs to be less than 0.05. As shown in \cref{tab:finetune}, DOV4MM is still valid in this more arduous scenario.

\begin{table}
\small
\addtolength{\tabcolsep}{1.7pt}
\begin{tabular}{ccccc}

\Xhline{1.0pt}
 $\mathcal{D}_{f}$& MIM Method& Model& w/o ft&  w/ ft \\ \hline

 \multirow{4}{*}{Food101} &\multirow{2}{*}{MAE} &ViT-B/16 &$10^{-5}$ &$10^{-5}$ \\
 & &ViT-L/16 &$10^{-5}$ &$10^{-4}$ \\ \cline{2-5}
 & \multirow{2}{*}{iBOT} &ViT-B/16 &$10^{-3}$ &$10^{-3}$ \\
 & &ViT-L/16 &$10^{-3}$ &$10^{-3}$ \\ \hline
 \multirow{4}{*}{Places365} &\multirow{2}{*}{MAE} &ViT-B/16 &$10^{-5}$ &$10^{-5}$ \\
 & &ViT-L/16 &$10^{-5}$ &$10^{-4}$ \\ \cline{2-5}
 & \multirow{2}{*}{iBOT} &ViT-B/16 &$10^{-3}$ &$10^{-3}$ \\
 & &ViT-L/16 &$10^{-3}$ &$10^{-3}$ \\
  \Xhline{1.0pt}
\end{tabular}
\vspace{-1em}
\caption{We report the $p$-values of $M_s$, whose pre-training dataset is ImageNet-1K, with or without fine-tuning (ft). $\mathcal{D}_{f}$ represents the fine-tuning dataset.}
\label{tab:finetune}
\end{table}

\paragraph{Adaptive attack.} We assume that the training objective of $M_s$ consists of two equally weighted components: (1) the proxy task, (2) a loss $L$ that minimizes the difference in reconstruction difficulty between seen and unseen samples. $M_s$ are ViT-B pre-trained with different subsets of ImageNet-1K ($\mathcal{D}_{pub}$). \cref{tab:adv} indicates that DOV4MM remains effective. This is because, although $L$ narrows the reconstruction gap between seen and unseen samples, the gap remains due to the proxy task.

\begin{table}[ht]
\centering
\small
\addtolength{\tabcolsep}{-3.2pt}
\begin{minipage}{0.23\textwidth}
\centering
\begin{tabular}{ccc}
\Xhline{1.0pt}
MIM Methods& \textbf{w/o} $L$& \textbf{w/} $L$ \\ \hline
 MAE &$10^{-9}$ &$10^{-9}$ \\
 CAE &$10^{-10}$ &$10^{-8}$ \\
  \Xhline{1.0pt}
\end{tabular}
\subcaption{$\mathcal{D}_{pub}$ is ImageNet-50.}
\end{minipage}
\hfill \hspace{2.5pt}
\begin{minipage}{0.23\textwidth}
\centering
\begin{tabular}{ccc}
\Xhline{1.0pt}
MIM Methods& \textbf{w/o} $L$& \textbf{w/} $L$ \\ \hline
 MAE &$10^{-5}$ &$10^{-5}$ \\
 CAE &$10^{-5}$ &$10^{-4}$ \\
\Xhline{1.0pt}
\end{tabular}
\subcaption{$\mathcal{D}_{pub}$ is ImageNet-100.}
\end{minipage}
\vspace{-1em}
\caption{$p$ (should $<$ 0.05) under different $\mathcal{D}_{pub}$.}
\label{tab:adv}
\end{table}

\section{Visualization Results}

\begin{figure}[ht]
    \centering
    \begin{subfigure}{0.49\textwidth}
        \centering
        \includegraphics[width=\textwidth, trim=12cm 6cm 3cm 3cm, clip]{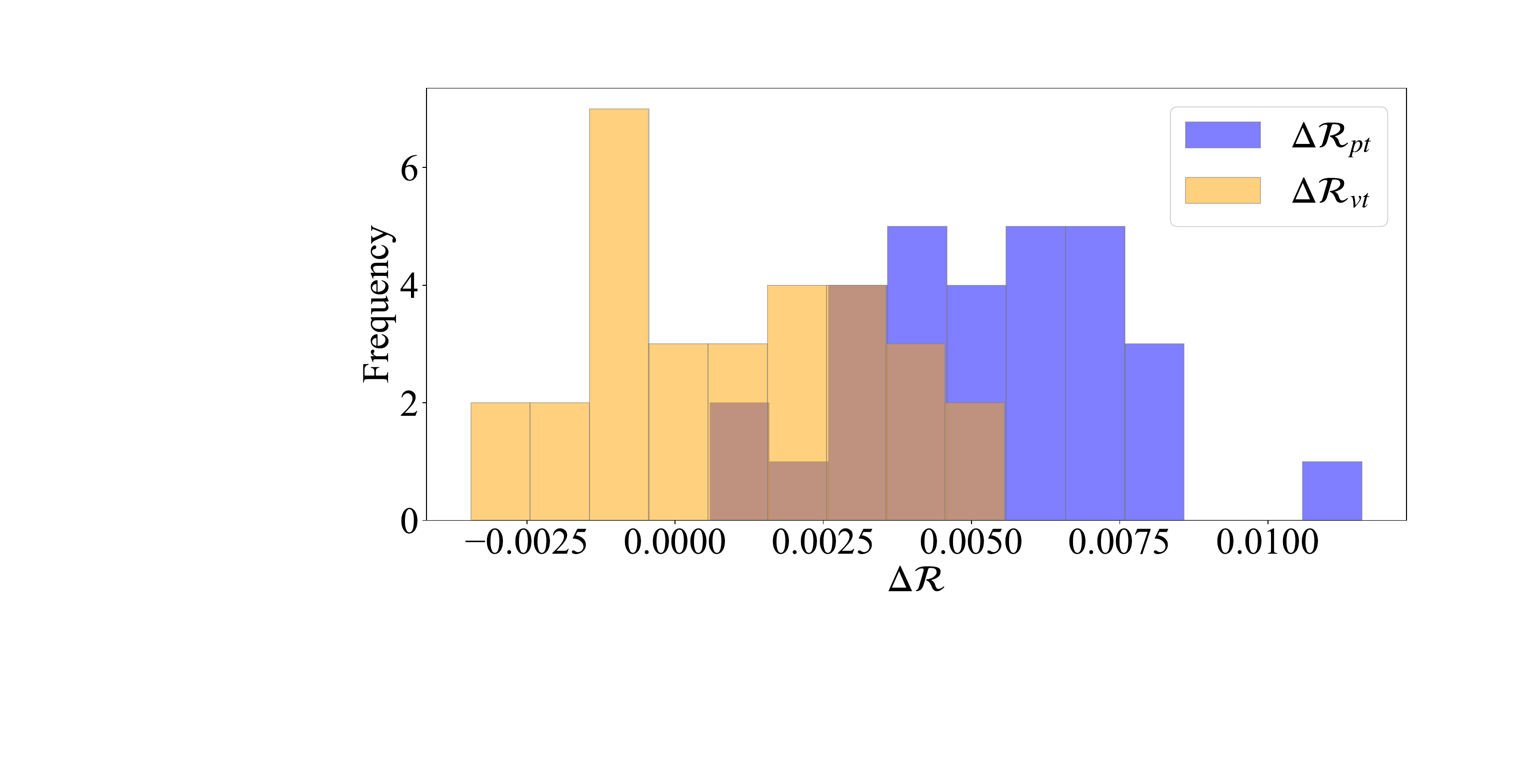}
        \caption{$\mathcal{D}_{pub}$ is ImageNet-50. ($M_s$ is illegal)}
        \label{fig:in50}
    \end{subfigure}
    \vspace{0.5em}
    \begin{subfigure}{0.49\textwidth}
        \centering
        \includegraphics[width=\textwidth, trim=10.5cm 6cm 3.5cm 2cm, clip]{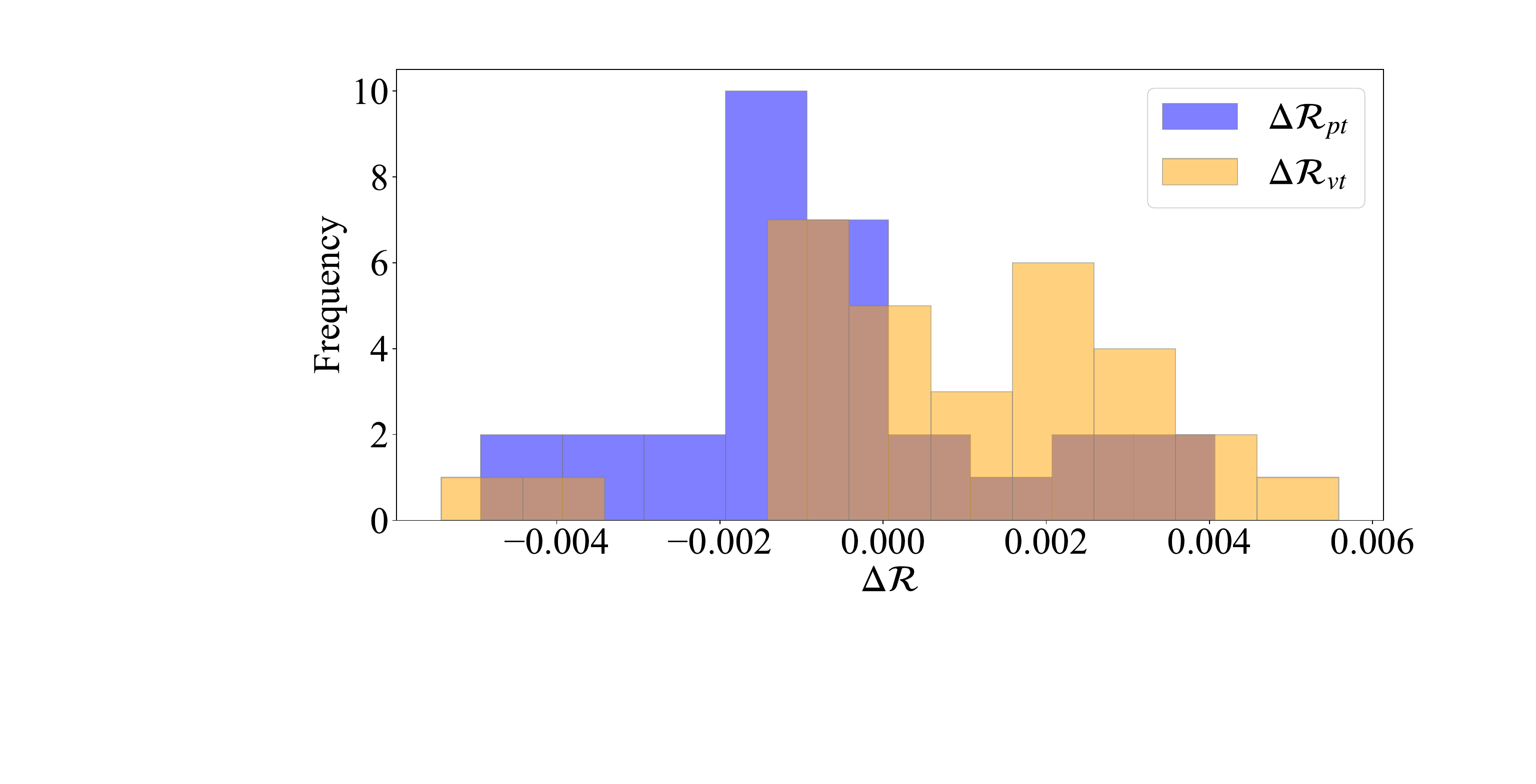}
        \caption{$\mathcal{D}_{pub}$ is Food101. ($M_s$ is legal)}
        \label{fig:food}
    \end{subfigure}
    \vspace{0.5em}
    \begin{subfigure}{0.49\textwidth}
        \centering
        \includegraphics[width=\textwidth, trim=12cm 6cm 3cm 3cm, clip]{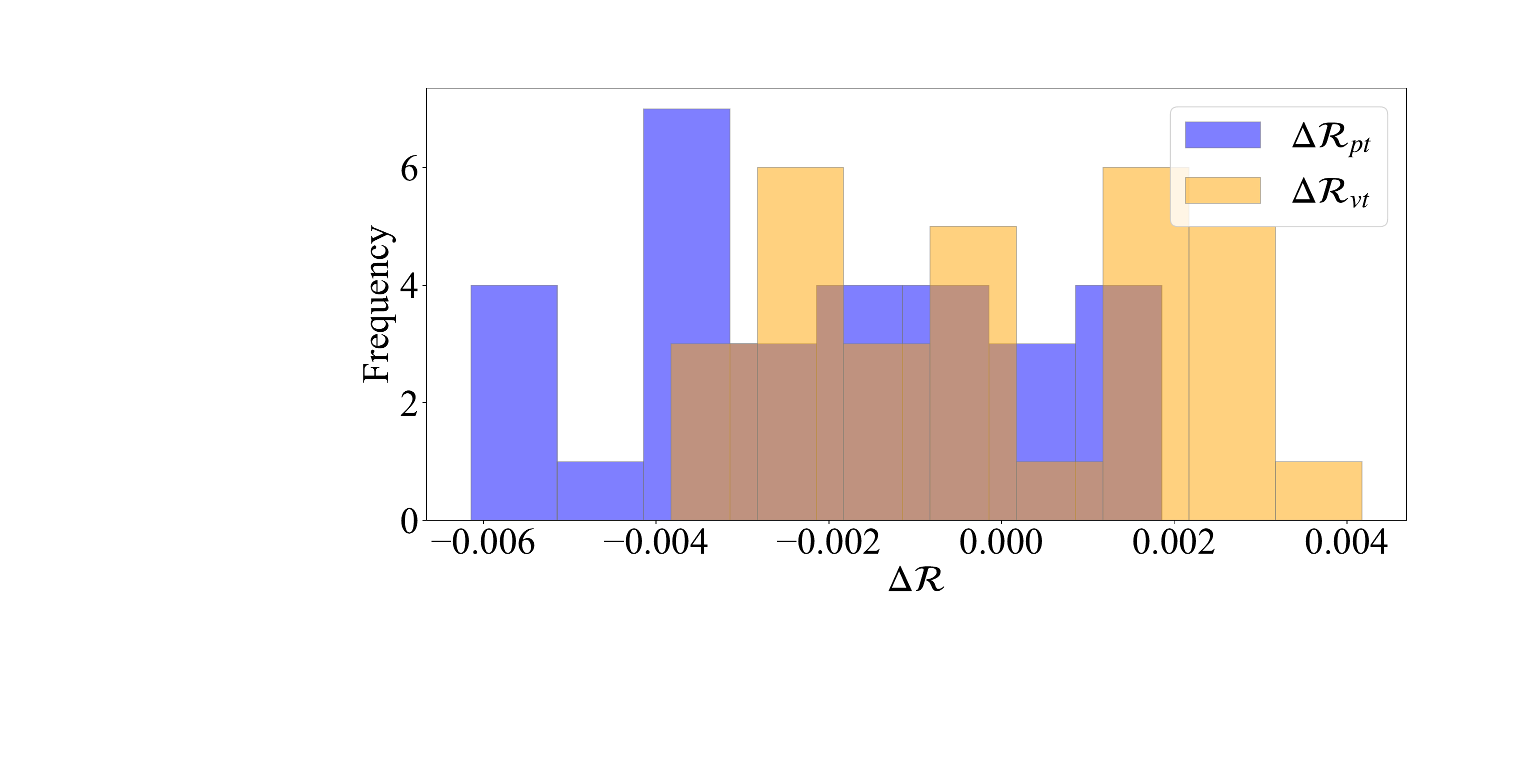}
        \caption{$\mathcal{D}_{pub}$ is COCO. ($M_s$ is legal)}
        \label{fig:coco}
    \end{subfigure}
    \vspace{0.5em}
    \begin{subfigure}{0.49\textwidth}
        \centering
        \includegraphics[width=\textwidth, trim=12cm 6cm 3cm 3cm, clip]{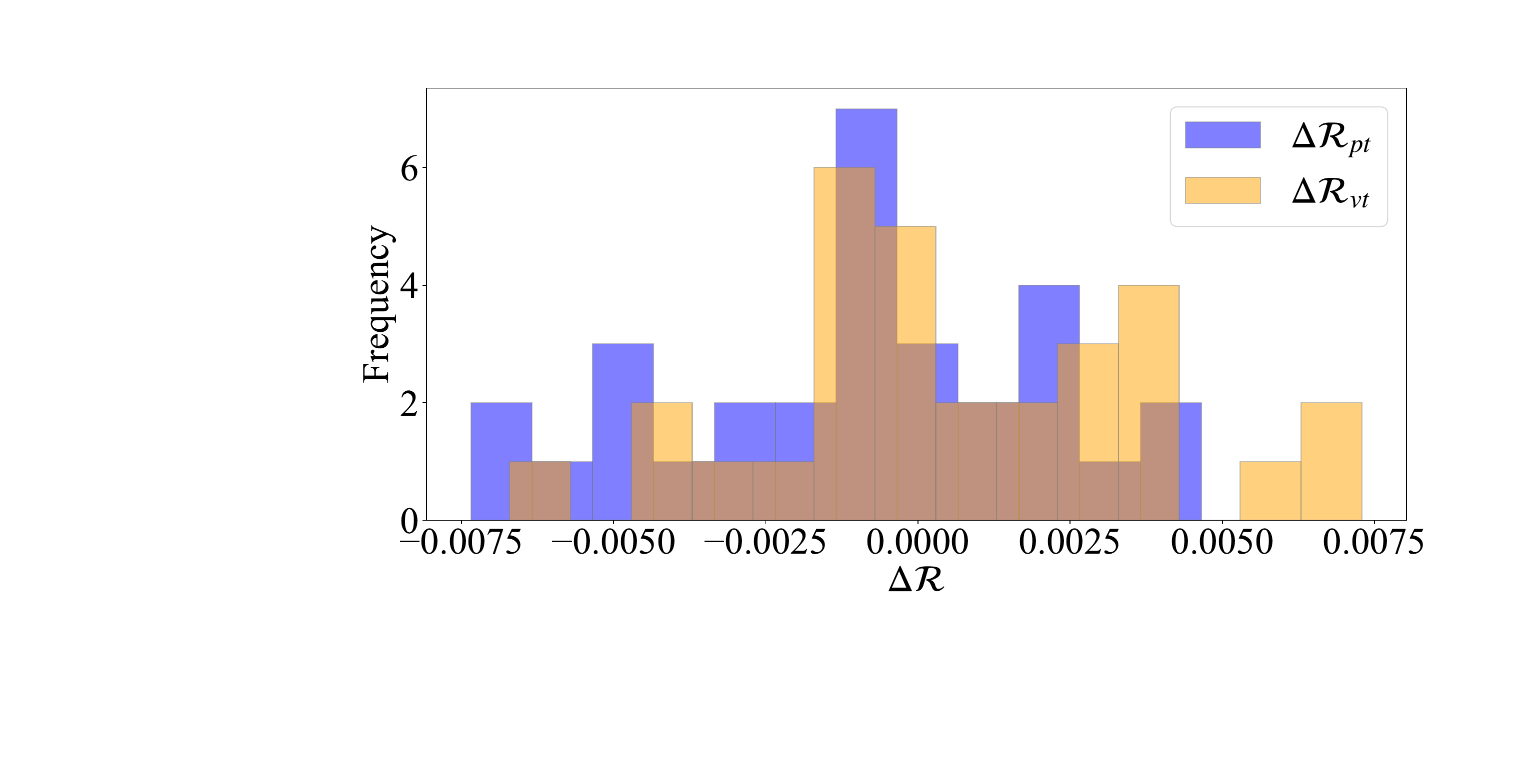}
        \caption{$\mathcal{D}_{pub}$ is Places365. ($M_s$ is legal)}
        \label{fig:places}
    \end{subfigure}
    \vspace{-2em}
    \caption{The distribution of relative embedding reconstruction difficulties \( \Delta \mathcal{R}_{vt} \) and \( \Delta \mathcal{R}_{pt} \). Note that when $\mathcal{D}_{pub}$ is ImageNet-50, $M_s$ is considered illegal, while in all other cases, it is legal.}
    \vspace{-1em}
    \label{fig:visual}
\end{figure}

We present the visualization results of DOV4MM on the ImageNet-50 pre-trained model. Specifically, \( \mathcal{D}_{pub} \) is set to ImageNet-50, Food101, COCO, or Places365, and the suspicious model is a ViT-L/16 pre-trained using MAE. The pre-training dataset of suspicious model is ImageNet-50. We computed the relative embedding reconstruction difficulties from suspicious models. Specifically, they include the relative reconstruction difficulty of embeddings between the training set \( \mathcal{D}_{t} \) and the validation set \( \mathcal{D}_{v} \) (\( \Delta \mathcal{R}_{vt} \)), as well as the relative reconstruction difficulty of embeddings between the training set \( \mathcal{D}_{t} \) and the test set \( \mathcal{D}_{pvt} \) (\( \Delta \mathcal{R}_{pt} \)), which are then visualized. When the suspicious model is pre-trained on \( \mathcal{D}_{pub} \), it is considered illegal, and \( \Delta \mathcal{R}_{pt} \) should be generally higher than \( \Delta \mathcal{R}_{vt} \). In contrast, if the suspicious model is legal, this phenomenon is not obvious.

As shown in \cref{fig:visual}, when the suspicious model is illegal, \( \Delta \mathcal{R}_{pt} \) is generally higher than \( \Delta \mathcal{R}_{vt} \). When the suspicious model is legal, there is no such relationship between the two relative embedding reconstruction difficulties. This observation is consistent with our previous findings.

\begin{table}
\centering
\small
\addtolength{\tabcolsep}{-3.0pt}
\begin{tabular}{cccccc}
\Xhline{1.0pt}
 \multirow{2}{*}{Model}&\multirow{2}{*}{MIM Method}&\multicolumn{4}{c}{\(\mathcal{D}_{pub}\)}\\ \cline{3-6}
 & & IN-50& Food101& COCO& Places365 \\ \hline
 \multirow{3}{*}{ViT-B/16}& MAE&  0.56&  0.99&  0.28&  0.46\\
 & CAE&  0.99&  1.00&  0.94&  0.90\\
 & iBOT&  0.09&  0.90&  0.41&  0.48\\ \hline
 \multirow{3}{*}{ViT-L/16}& MAE&  0.31&  0.91&  0.40&  0.59\\
 & CAE&  0.99&  1.00&  0.66&  0.96\\
 & iBOT&  0.31&  0.91&  0.41&  0.59\\
\Xhline{1.0pt}
\end{tabular}
\vspace{-1em}
\caption{The results (\(p\)-values) of DI4SSL on ImageNet-50. ``IN-50'' represents ImageNet-50.}
\label{tab:di4ssl_in50}
\vspace{-1em}
\end{table}

\begin{table}
\centering
\small
\addtolength{\tabcolsep}{-3.0pt}
\begin{tabular}{cccccc}
\Xhline{1.0pt}
 \multirow{2}{*}{Model}&\multirow{2}{*}{MIM Method}&\multicolumn{4}{c}{\(\mathcal{D}_{pub}\)}\\ \cline{3-6}
 & & IN-50& Food101& COCO& Places365 \\ \hline
 \multirow{3}{*}{ViT-B/16}& MAE&  $10^{-155}$&  $10^{-113}$&  $10^{-134}$&  $10^{-140}$\\
 & CAE&  $10^{-284}$&  $10^{-148}$&  $10^{-246}$&  $10^{-255}$\\
 & iBOT&  $10^{-155}$&  $10^{-113}$&  $10^{-134}$&  $10^{-140}$\\ \hline
 \multirow{3}{*}{ViT-L/16}& MAE&  $10^{-160}$&  $10^{-122}$&  $10^{-140}$&  $10^{-152}$\\
 & CAE&  $10^{-199}$&  $10^{-148}$&  $10^{-171}$&  $10^{-151}$\\
 & iBOT&  $10^{-160}$&  $10^{-122}$&  $10^{-140}$&  $10^{-152}$\\
\Xhline{1.0pt}
\end{tabular}
\vspace{-1em}
\caption{The results (\(p\)-values) of CTRL on ImageNet-50. ``IN-50'' represents ImageNet-50.}
\label{tab:ctrl_in50}
\vspace{-1em}
\end{table}

\begin{table}
\centering
\small
\addtolength{\tabcolsep}{-3.0pt}
\begin{tabular}{cccccc}
\Xhline{1.0pt}
 \multirow{2}{*}{Model}&\multirow{2}{*}{MIM Method}&\multicolumn{4}{c}{\(\mathcal{D}_{pub}\)}\\ \cline{3-6}
 & & IN-50& Food101& COCO& Places365 \\ \hline
 \multirow{3}{*}{ViT-B/16}& MAE&  0.28&  0.12&  0.50&  0.15\\
 & CAE&  0.06&  0.48&  0.25&  $10^{-3}$\\
 & iBOT&  0.28&  0.11&  0.50&  0.15\\ \hline
 \multirow{3}{*}{ViT-L/16}& MAE&  0.12&  0.06&  0.56&  0.03\\
 & CAE&  0.16&  0.26&  0.34&  0.09\\
 & iBOT&  0.12&  0.06&  0.56&  0.01\\
\Xhline{1.0pt}
\end{tabular}
\vspace{-1em}
\caption{The results (\(p\)-values) of PartCrop on ImageNet-50. ``IN-50'' represents ImageNet-50.}
\label{tab:partcrop_in50}
\vspace{-1em}
\end{table}

\begin{table}
\centering
\small
\addtolength{\tabcolsep}{-3.0pt}
\begin{tabular}{cccccc}
\Xhline{1.0pt}
 \multirow{2}{*}{Model}&\multirow{2}{*}{MIM Method}&\multicolumn{4}{c}{\(\mathcal{D}_{pub}\)}\\ \cline{3-6}
 & & IN-50& Food101& COCO& Places365 \\ \hline
 \multirow{3}{*}{ViT-B/16}& MAE&  $10^{-9}$&  0.92&  0.99&  0.99\\
 & CAE&  $10^{-10}$&  0.91&  0.99&  0.90\\
 & iBOT&  $10^{-4}$&  0.99&  0.99&  0.97\\ \hline
 \multirow{3}{*}{ViT-L/16}& MAE&  $10^{-12}$&  0.99&  0.99&  0.98\\
 & CAE&  $10^{-9}$&  0.96&  0.99&  0.92\\
 & iBOT&  $10^{-6}$&  0.99&  0.99&  0.89\\
\Xhline{1.0pt}
\end{tabular}
\vspace{-1em}
\caption{The results (\(p\)-values) of DOV4MM on ImageNet-50. ``IN-50'' represents ImageNet-50.}
\label{tab:dov4mm_in50}
\vspace{-1em}
\end{table}

\begin{table}
\centering
\small
\addtolength{\tabcolsep}{-3.0pt}
\begin{tabular}{cccccc}
\Xhline{1.0pt}
 \multirow{2}{*}{Model}&\multirow{2}{*}{MIM Method}&\multicolumn{4}{c}{\(\mathcal{D}_{pub}\)}\\ \cline{3-6}
 & & IN-100& Food101& COCO& Places365 \\ \hline
 \multirow{3}{*}{ViT-B/16}& MAE&  0.68&  0.90&  0.41&  0.46\\
 & CAE&  0.99&  0.99&  0.98&  0.53\\
 & iBOT&  0.79&  0.91&  0.41&  0.46\\ \hline
 \multirow{3}{*}{ViT-L/16}& MAE&  0.83&  0.83&  0.41&  0.38\\
 & CAE&  0.99&  1.00&  0.90&  0.47\\
 & iBOT&  0.88&  0.79&  0.40&  0.59\\
\Xhline{1.0pt}
\end{tabular}
\vspace{-1em}
\caption{The results (\(p\)-values) of DI4SSL on ImageNet-100. ``IN-100'' represents ImageNet-100.}
\label{tab:di4ssl_in100}
\vspace{-1em}
\end{table}

\begin{table}
\centering
\small
\addtolength{\tabcolsep}{-3.0pt}
\begin{tabular}{cccccc}
\Xhline{1.0pt}
 \multirow{2}{*}{Model}&\multirow{2}{*}{MIM Method}&\multicolumn{4}{c}{\(\mathcal{D}_{pub}\)}\\ \cline{3-6}
 & & IN-100& Food101& COCO& Places365 \\ \hline
 \multirow{3}{*}{ViT-B/16}& MAE&  $10^{-150}$&  $10^{-113}$&  $10^{-134}$&  $10^{-140}$\\
 & CAE&  $10^{-236}$&  $10^{-129}$&  $10^{-225}$&  $10^{-235}$\\
 & iBOT&  $10^{-150}$&  $10^{-113}$&  $10^{-134}$&  $10^{-140}$\\ \hline
 \multirow{3}{*}{ViT-L/16}& MAE&  $10^{-158}$&  $10^{-122}$&  $10^{-140}$&  $10^{-152}$\\
 & CAE&  $10^{-173}$&  $10^{-120}$&  $10^{-165}$&  $10^{-144}$\\
 & iBOT&  $10^{-158}$&  $10^{-122}$&  $10^{-140}$&  $10^{-152}$\\
\Xhline{1.0pt}
\end{tabular}
\vspace{-1em}
\caption{The results (\(p\)-values) of CTRL on ImageNet-100. ``IN-100'' represents ImageNet-100.}
\label{tab:ctrl_in100}
\vspace{-1em}
\end{table}

\begin{table}
\centering
\small
\addtolength{\tabcolsep}{-3.0pt}
\begin{tabular}{cccccc}
\Xhline{1.0pt}
 \multirow{2}{*}{Model}&\multirow{2}{*}{MIM Method}&\multicolumn{4}{c}{\(\mathcal{D}_{pub}\)}\\ \cline{3-6}
 & & IN-100& Food101& COCO& Places365 \\ \hline
 \multirow{3}{*}{ViT-B/16}& MAE&  0.42&  0.11&  0.50&  0.15\\
 & CAE&  0.10&  0.27&  0.28&  $10^{-3}$\\
 & iBOT&  0.42&  0.11&  0.50&  0.15\\ \hline
 \multirow{3}{*}{ViT-L/16}& MAE&  0.33&  0.06&  0.56&  0.03\\
 & CAE&  0.19&  0.33&  0.38&  0.01\\
 & iBOT&  0.33&  0.06&  0.56&  0.03\\
\Xhline{1.0pt}
\end{tabular}
\vspace{-1em}
\caption{The results (\(p\)-values) of PartCrop on ImageNet-100. ``IN-100'' represents ImageNet-100.}
\label{tab:partcrop_in100}
\vspace{-1em}
\end{table}

\begin{table}
\centering
\small
\addtolength{\tabcolsep}{-3.0pt}
\begin{tabular}{cccccc}
\Xhline{1.0pt}
 \multirow{2}{*}{Model}&\multirow{2}{*}{MIM Method}&\multicolumn{4}{c}{\(\mathcal{D}_{pub}\)}\\ \cline{3-6}
 & & IN-100& Food101& COCO& Places365 \\ \hline
 \multirow{3}{*}{ViT-B/16}& MAE&  $10^{-5}$&  0.98&  0.99&  0.99\\
 & CAE&  $10^{-5}$&  0.96&  0.99&  0.92\\
 & iBOT&  $10^{-4}$&  0.99&  0.96&  0.97\\ \hline
 \multirow{3}{*}{ViT-L/16}& MAE&  $10^{-6}$&  0.99&  0.99&  0.98\\
 & CAE&  $10^{-4}$&  0.97&  0.99&  0.93\\
 & iBOT&  $10^{-4}$&  0.99&  0.99&  0.87\\
\Xhline{1.0pt}
\end{tabular}
\vspace{-1em}
\caption{The results (\(p\)-values) of DOV4MM on ImageNet-100. ``IN-100'' represents ImageNet-100.}
\label{tab:dov4mm_in100}
\vspace{-1em}
\end{table}

\begin{table}
\centering
\small
\addtolength{\tabcolsep}{-3.0pt}
\begin{tabular}{cccc}
\Xhline{1.0pt}
 \multirow{2}{*}{MLM Model}&\multicolumn{3}{c}{\(\mathcal{D}_{pub}\)}\\ \cline{2-4}
 & Wiki-50k& CC-News& MiniPile \\ \hline
 BERT\textsubscript{Base}&  $10^{-16}$&  0.53&  0.99 \\
 BERT\textsubscript{Large}&  $10^{-13}$&  0.46&  0.93 \\
 RoBERTa\textsubscript{Base}&  $10^{-18}$&  0.71&  0.57 \\
 RoBERTa\textsubscript{Large}&  $10^{-9}$&  0.68&  0.74 \\
 ALBERT\textsubscript{Base}&  $10^{-34}$&  0.43&  0.54 \\
 XLM-R\textsubscript{Base}&  $10^{-18}$&  0.32&  0.13 \\
\Xhline{1.0pt}
\end{tabular}
\vspace{-1em}
\caption{The results (\(p\)-values) of DOV4MM on WikiText-103-50k. ``Wiki-50k'' represents WikiText-103-50k.}
\label{tab:wiki50k}
\vspace{-1em}
\end{table}

\begin{table}
\centering
\small
\addtolength{\tabcolsep}{-3.0pt}
\begin{tabular}{cccc}
\Xhline{1.0pt}
 \multirow{2}{*}{MLM Model}&\multicolumn{3}{c}{\(\mathcal{D}_{pub}\)}\\ \cline{2-4}
 & Wiki-100k& CC-News& MiniPile \\ \hline
 BERT\textsubscript{Base}&  $10^{-10}$&  0.72&  0.97 \\
 BERT\textsubscript{Large}&  $10^{-10}$&  0.69&  0.82 \\
 RoBERTa\textsubscript{Base}&  $10^{-3}$&  0.48&  0.58 \\
 RoBERTa\textsubscript{Large}&  $10^{-12}$&  0.47&  0.55 \\
 ALBERT\textsubscript{Base}&  $10^{-9}$&  0.10&  0.54 \\
 XLM-R\textsubscript{Base}&  $10^{-7}$&  0.20&  0.08 \\
\Xhline{1.0pt}
\end{tabular}
\vspace{-1em}
\caption{The results (\(p\)-values) of DOV4MM on WikiText-103-100k. ``Wiki-100k'' represents WikiText-103-100k.}
\label{tab:wiki100k}
\vspace{-1em}
\end{table}

\section{The Specific \texorpdfstring{$p$}{Lg}-values in \texorpdfstring{\cref{fig:main_exp}}{Lg} and \texorpdfstring{\cref{fig:nlp_exp}}{Lg}}
Here, we present the specific results of different methods on various datasets (as shown in \cref{fig:main_exp} of the paper), as shown in \cref{tab:di4ssl_in50} - \cref{tab:dov4mm_in100}. Note that \(p\) should be less than 0.05 in the illegal scenarios and greater than 0.05 in the legal scenarios.

We also present the specific results of different methods on various datasets (as shown in \cref{fig:nlp_exp} of the paper), as shown in \cref{tab:wiki50k} and \cref{tab:wiki100k}. Note that \(p\) should be less than 0.05 in the illegal scenarios and greater than 0.05 in the legal scenarios.


\end{document}